%% file: log_2024.tex
\documentclass{article}
\usepackage[preprint]{log_2024}			% for preprint version

\usepackage{booktabs}						% professional-quality tables
\usepackage{multirow}						% tabular cells spanning multiple rows
\usepackage{amsfonts}						% blackboard math symbols
\usepackage{graphicx}						% figures
\usepackage{duckuments}						% sample images

\usepackage[numbers,compress,sort]{natbib}	% for numerical citations
\usepackage{paralist}
\usepackage[linesnumbered,lined,boxed,commentsnumbered]{algorithm2e}
\SetKw{KwCal}{calculate}
\usepackage{array}

\usepackage{caption}
\usepackage{subcaption}
\usepackage{pifont}

\usepackage{xcolor}
\usepackage{colortbl}

\newcommand{\method}{\textsc{SGDiff}\xspace}

\DeclareMathOperator*{\argmin}{arg\,min}

\title{Sub-graph Based Diffusion Model for Link Prediction}

\author[H. Li et al.]{%
Hang Li\\
Michigan State University\\
\email{lihang4@msu.edu}
\And
Wei Jin\\
Emory University\\
\email{wei.jin@emory.edu}
\And
Geri Skenderi\\
Bocconi University\\
\email{geri.skenderi@unibocconi.it}
\And
Harry Shomer\\
Michigan State University\\
\email{shomerha@msu.edu}
\And
Wenzhuo Tang\\
Michigan State University\\
\email{tangwen2@msu.edu}
\And
Wenqi Fan\\
The Hong Kong Polytechnic University\\
\email{wenqi.fan@polyu.edu.hk}
\And
Jiliang Tang\\
Michigan State University\\
\email{tangjili@msu.edu}
}

\begin{document}

\maketitle

\begin{abstract}

\input{abstract}

\end{abstract}

\section{Introduction}

\input{introduce_old}

\section{Related Work}
\input{preliminaries}

\section{Method}
\input{methodology}

\section{Experiments}
\input{experiment}

\section{Conclusion}
\input{conclusion}

% For natbib users:
\bibliographystyle{unsrtnat}
\bibliography{reference}
% For bibLaTeX users:
% \printbibliography

\appendix

\input{appendix}

\end{document}

%% file: abstract.tex
Denoising Diffusion Probabilistic Models (DDPMs) represent a contemporary class of generative models with exceptional qualities in both synthesis and maximizing the data likelihood. These models work by traversing a forward Markov Chain where data is perturbed, followed by a reverse process where a neural network learns to undo the perturbations and recover the original data. There have been increasing efforts exploring the applications of DDPMs in the graph domain. However, most of them have focused on the generative perspective. In this paper, we aim to build a novel generative model for link prediction. In particular, we treat link prediction between a pair of nodes as a conditional likelihood estimation of its enclosing sub-graph. With a dedicated design to decompose the likelihood estimation process via the Bayesian formula, we are able to separate the estimation of sub-graph structure and its node features. Such designs allow our model to simultaneously enjoy the advantages of  inductive learning and the strong generalization capability. Remarkably, comprehensive experiments across various datasets validate that our proposed method presents numerous advantages: (1) transferability across datasets without retraining, (2) promising generalization on limited training data, and (3) robustness against graph adversarial attacks.

%% file: introduce_old.tex
Graphs are ubiquitous data structures, with applications that span from social networks~\cite{tang2010graph,qiu2018deepinf,wang2019mcne} to cutting-edge scientific research~\cite{liu2018constrained,wang2022molecular,wang2021leverage,wen2022graph}. Link prediction, as one of the most fundamental tasks on graphs, plays an indispensable role in various graph applications in web-related scientific researches such as e-commence recommendations~\cite{koren2009matrix,fan2022graph}, social network analysis~\cite{hasan2011survey}, and network security predictions~\cite{pope2019automated}. With the recent rise of graph neural networks (GNNs), a variety of GNN-based methods have been developed, tremendously advancing the performance of link prediction~\cite{zhang2018link, chamberlain2022graph, wang2023neural}. 

In GNN-based link prediction, two main families of techniques have been proposed: discriminative methods~\cite{zhang2018link,zhu2021neural,chamberlain2022graph} and generative methods~\cite{kipf2016variational,chen2019generative}. 
While discriminative methods are more popular, the utilization of traditional generative models (e.g., VGAEs~\cite{kipf2016variational}) remains rather limited. The exploration of recent generative approaches, e.g., diffusion models~\cite{ho2020denoising,song2020score}, is even less prevalent. Nevertheless, generative models are well-known for their advantages in generalization and robustness~\cite{ng2001discriminative,grathwohl2019your, chen2023robust}, particularly in scenarios with limited labeled data or under adversarial attacks, where they often outperform their discriminative counterparts. In fact, adopting generative models for discriminative tasks has recently gained increasing attention in various domains such as computer vision~\cite{ravuri2019classification} and natural language processing~\cite{liu2023pre}. For example, the GPT series models~\cite{brown2020language,ouyang2022training} have fully demonstrated the potential of such generative models by showcasing exceptional generalization abilities. This involves solving not only text generation tasks but also numerous classification problems~\cite{radford2019language,bubeck2023sparks}. Meanwhile, diffusion models~\cite{ho2020denoising,song2020score} have exhibited remarkable competence in discriminative tasks in the image domain including image classification~\cite{zimmermann2021score,li2023your} and image segmentation~\cite{amit2021segdiff}. 
Given these developments, we are motivated to explore a generative approach for a fundamental discriminative graph problem, i.e., link prediction. Such efforts not only have a great potential to enhance the generalization and robustness of link prediction but also can inspire the application of generative models in other standard graph learning tasks such as node classification and graph classification.

However, developing generative models for the link prediction problem faces unique challenges. First, the size of the graph adjacency matrix increases with the increase of node size. Thus, modeling the whole graph structure with the generative model in one-shot ~\cite{kipf2016variational} leads to an excessive memory footprint for large graphs. One possible approach to circumvent this issue is to rely on autoregressive modeling ~\cite{li2018learning}. However, due to its low efficiency and high variance~\cite{bu2023let}, the autoregressive generation model has not been used for graph learning tasks. To tackle this challenge, we propose a sub-graph based diffusion framework \method, that take advantage of the recent success of sub-graph based GNNs~\cite{zhang2018link}. By using only the sub-graph, the size of the sub-graph adjacency matrix is relatively much smaller than the whole graph. Second, we need to contend with the node features in addition to graph structure. Prior generative works~\cite{kipf2016variational} only use node features as inputs to reconstruct the graph structure. This design causes the generative model to loses the capability to transfer between different datasets, as nodes features of different datasets are usually incompatible with each other. This limits the generalization capabilities of such generative models, leaving this area under-explored. These challenges motivate us in the design of a new framework -- \method. \method uses Bayesian theorem to decompose the generation of graph structure and node features into consequential steps, thereby helping achieve success in its cross dataset transfer capabilities.

%% file: preliminaries.tex
% In line with the focus of our work, we briefly describe related work on sub-graph based link prediction and likelihood estimation of diffusion models.

% \hang{Give the background info of using sub-graph classification framework for link prediction task, presenting some theories covered by SEAL and introduce briefly about the SEAL algorithm. Please also mention the disadvantages of some existing method like unfit for inductive settings and for the ones with inductive settings they may behave so well due to the over-fitting issue. Besides, some advanced heuristic features may needed which make this goal non-trivial. One thing may need to be clear, whether we claim we are better at cross-domain setting or induction setting. Using the inductive setting will be risky as we do not have the experiment result on knowledge graph completion: FB15k-237 and WN18RR. The read-out function could also be an defect of sub-graph based methods.}
% \footnote{GNNs are a broader category than MPNNs. Since the majority of GNNs used in practice are of the message passing type, we use both terms interchangeably.}
% \vspace{-0.1cm}
\subsection{Sub-graph Based Link Prediction}
Due to the limitation of traditional Message Passing Neural Network (MPNN) in capturing the pairwise relations between two individual target nodes, vanilla GNNs often  struggle with link prediction problems~\citep{zhang2021labeling}. To solve this issue, manual feature enhanced models (MFEMs) like NBFNet~\citep{zhu2021neural}, NCNC~\citep{wang2023neural} and BUDDY~\citep{chamberlain2022graph} proposed a variety of methods trying to fuse the complementary structure information, e.g., heuristic features, with the message passing neural networks. On the other hand, sub-graph GNNs (SGNNs) like SEAL~\citep{zhang2018link} and SUREL~\citep{yin2022algorithm} transform link prediction into a binary sub-graph classification task and attempts to learn data-driven link prediction heuristics. Compared with fusion-based algorithms, SGNNs do not require complicated heuristic feature fusion designs and have better generous capability to different datasets. Besides, since they use sub-graphs as the sample unit, SGNNs are more flexible to inductive scenarios \citep{teru2020inductive}. 

Next we give a formal statement about SGNNs. For a pair of nodes $u$, $v$ and its enclosing sub-graph $G_{uv}$, SGNNs produce sub-graph representation $Y_{u,v}$ with GNNs and desired read-out functions $\mathcal{R}$. With the classifier $\mathcal{C}$, the sub-graph representation $Y_{u,v}$ is expected to classified as one if an edge $(u,v)$ exists and zero otherwise. Commonly, node features are augmented with structural features to resolve the automorphic node problem~\citep{zhang2021labeling}. The global heuristics can be well approximated from sub-graphs that are augmented with structural features with an approximation error that decays exponentially with the number of hops taken to construct the sub-graph~\citep{zhang2018link}. By incorporating the idea of SGNNs, we aim to solve the memory print challenge for generative models on graph learning problems.

% to take its advantages in inductive learning and get rid of memory print problem used 
% get rid of prior whole graph modeling style and 
% take its advantages in inductive  
% In \method, we incorporate the similar idea of SGNNs, and use generative model to estimate the connection likelihood score of the observed enclosing sub-graph $G_{uv}$.

% \subsection{Likelihood Estimation of Diffusion Model}

\subsection{Likelihood Estimation of Diffusion Models}\label{sec:likelihood}
% \vspace{-0.1cm}
% \hang{Give some information about diffusion models (generative perspective, give some references) and cover diffusion model's capability in estimating the likelihood of given samples.}
% \geri{I think there is some confusion here. DDPMs can be considered explicit models since they have a likelihood-based formulation. On the other hand, score-based models (note the citation to Song et al in this paragraph) are implicit models, since they estimate the gradient of the data distribution. So, a particular formulation of score-based models with variance preserving diffusion and an ELBO objective for MLE (aka DDPMs) can be considered explicit, but score-based models in general are not. This is important to not create confusion amongst reviewers who might be very cautious towards these details.}
Diffusion models~\citep{sohl2015deep,ho2020denoising} are a contemporary class of generative models. Through an iterative noising (forward) and denoising (reverse) Markov chain, diffusion models aim to learn the distribution of data in an explicit way ~\citep{sohl2015deep}. Diffusion models enjoy the benefit of having a likelihood-based objective like VAEs \citep{kingma2013auto} as well as high visual sample quality like GANs~\citep{goodfellow2020generative} even on high variability datasets. Recent advances in this area have also shown amazing results in text-to-image generation~\citep{ramesh2022hierarchical,saharia2022photorealistic,rombach2022high}, audio synthesis~\citep{kong2020diffwave,liu2023audioldm} and text-to-3d content creation~\citep{poole2022dreamfusion,lin2023magic3d}. 
Despite being powerful generative models, diffusion models has also been recently recognized as valid generative classifiers~\citep{clark2023text,mukhopadhyay2023diffusion}. As using the variational lower bound (VLB) of the log-likelihood as the object function, a well-trained diffusion models could provide accurate estimations to the probability of samples within the data distribution~\citep{li2023your}. Furthermore, by incorporating class information as the condition input during the training, the diffusion model can be used to compute class-conditional likelihoods $p_{\theta}(\mathbf{x}|\mathbf{y})$. Then, by selecting an appropriate prior distribution $p(\mathbf{y})$ and applying Bayes' theorem, predicted class probabilities $p_{\theta}(\mathbf{y}|\mathbf{x})$ can be easily calculated. Compared to discriminative models, generative classifiers have been shown to generalize better, be more robust, and be better calibrated \citep{grathwohl2019your, chen2023robust}. 
In this work, we seek to develop diffusion models for solving discriminative graph problems.

% \wei{add one sentence to describe our work or our goal..}

% \begin{itquote}
%     Given a graph $\mathcal{G}$
% \end{itquote}

% . And the notification 

%% file: methodology.tex
% \hang{Introduce the overall Bayesian formula decomposition for the task: $\mathcal{P}(Y|X,A) \propto \mathcal{P}(X|A,Y) * \mathcal{P}(A|Y) * \mathcal{P}(Y)$, and call out the diffusion model as the estimator for our the likelihood scores: $\mathcal{P}(X|A,Y)$(feature score) and $\mathcal{P}(A|Y)$ (structure score). Also mentions the motivation why we use diffusion model as the estimator, e.g., some strength of diffusion model and give some examples in image area using the likelihood score estimated by diffusion for some non-generative tasks.}
% \jt{You can start the section with introducing the organziation structure of the section. I think we should add one subsection about "Design Overview"}

Although there are recent works on applying diffusion models for problems on graphs~\citep{vignac2022digress,fan2023generative}, most of them focus on its generative perspective. The usage of the likelihood score of diffusion models to graph problems is relatively underexplored. Therefore, in this work, we take one of the most fundamental problems on graphs (i.e., link prediction) as an example to demonstrate the effectiveness of diffusion models with SGNNs for problems with graph data. Note that our algorithm can be easily extended to other graph problems like node or graph classification and we leave it as one future work. Next, we will first define notations and introduce an overview design of our algorithm \method. then, we present details about the link likelihood score estimation with the combination of structure and feature diffusion models.

\subsection{Notations}

In the following, we formally define the notations used in this work.  Let $G=(\mathcal{V},\mathcal{E})$ be an undirected graph where $\mathcal{V}$ and $\mathcal{E}$ denote the sets of $n$ nodes (vertices) $\mathcal{V}$ and $e$ links (edges), respectively. Let $S=(\mathcal{V}_S\subseteq\mathcal{V},\mathcal{E}_S\subseteq\mathcal{E})$ be a node-induced sub-graph of $G$ statisfying $(u,v)\in\mathcal{E}_S$ iff $(u,v)\in\mathcal{E}$ for any $u,v\in\mathcal{V}_S$. We use $S^k_{uv}=(\mathcal{V}_{uv},\mathcal{E}_{uv})$ to denote a $k$-hop sub-graph enclosing the link $(u,v)$, where $\mathcal{V}_{uv}$ is the union of the $k$-hop neighbors of $u$ and $v$ and $\mathcal{E}_{uv}$ is the union of the links that can be reached by a $k$-hop walk originating at $u$ and $v$. The given features of nodes $\mathcal{V}_{uv}$ are represented by $\mathbf{X}_{uv}$ and the adjacency matrix of $S^k_{uv}$ is $\mathbf{A}_{uv}$. The probability of link $(u,v)$ existing is indicated by $p(y_{uv}=1)$ and our goal is to estimate $p(y_{uv}=1 | \mathbf{X}_{uv}, \mathbf{A}_{uv})$ with likelihood score generated by diffusion models. As following parts process each sub-graph with the same process, we will omit the subscripts $u$ and $v$ for convenience.

\subsection{Design Overview}
\label{sec:overview}
% The recent success of SGNNs demonstrated the effectiveness of local-wised structure information. By incorporating the sub-graph modeling idea into our generative modeling framework, we will not only inherit the good inductive learning properties of SGNNs methods, but also solve the scalability issue in existing generative-based works. Based on that, we now formulate the link prediction probelm as  

% \hang{Here we need to add few words to introduce why we need enclosing subgraph, e.g., improve the scalability of generative model et al.}

As mentioned in Section~\ref{sec:likelihood}, by using a prior distribution $p(y)$, the categorical probability $p(y|x)$ can be estimated by applying Bayesian theorem over the class-conditional likelihood $p(x|y)$. However, unlike applying diffusion models to a single input like image or text, performing diffusion on graph data involves two different but correlated inputs, i.e., node feature ($\mathbf{X}$) and adjacency matrix ($\mathbf{A}$). Therefore, we need dedicated designs to decompose $p(y|\mathbf{A},\mathbf{X})$. To break the generalization limitation of existing generative approaches and take advantage of the inductive learning capability of SGNNs, we propose the following formulation:

% decide to 

% There major issue with prior graph generative methods is that the input node features $X_$ are incompatible across different datasets  

% generation is 

% \jt{I think we need more intuitions about why we want to decompose like the following equation. meanwhile we may have more options to do the decomposion, why the followign one and we need to justify our choices and its advntages} 

\begin{figure}[!btph]
\begin{center}
%\framebox[4.0in]{$\;$}
\includegraphics[width=\textwidth]{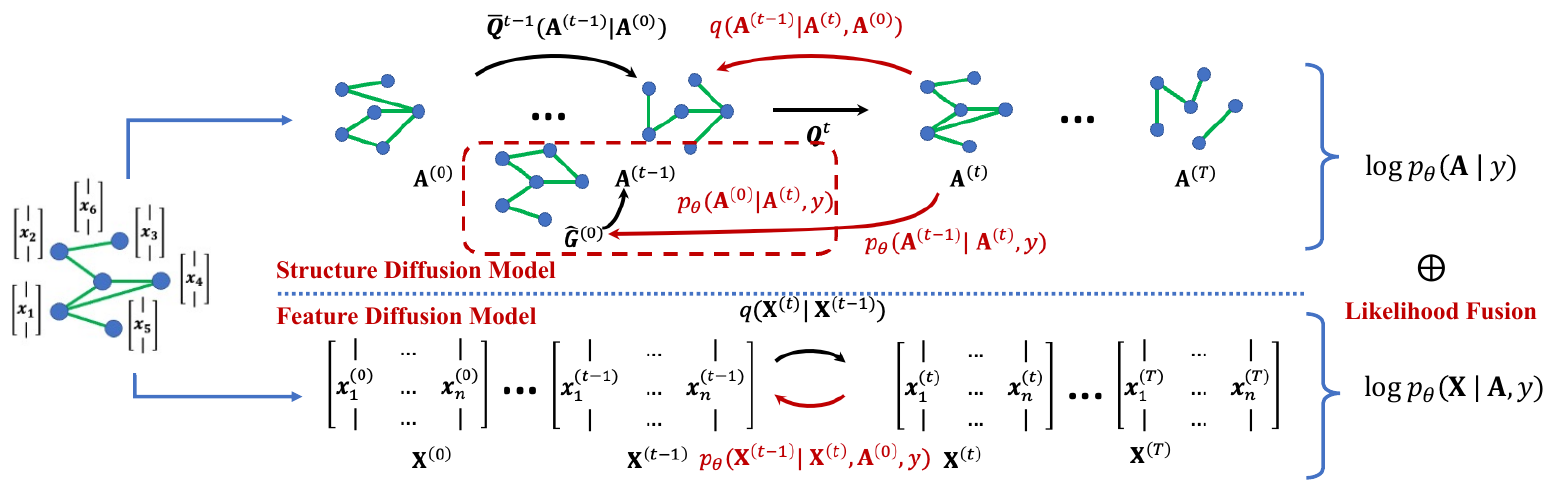}
\end{center}
\vspace{-0.3cm}
\caption{An overview of our proposed framework. $\mathbf{Q}^t$ and $q$ are diffusion kernels for structure and feature diffusion models, respectively. The calculation of log-likelihood scores $\log P_{\theta}(\mathbf{A}|y)$ and $\log P_{\phi}(\mathbf{X}|\mathbf{A},y)$ is based on fitted denoising models, $p_{\phi}(\mathbf{A}^{(0)}|\mathbf{A}^{(t)},y)$ and $p_{\epsilon}(\mathbf{X}^{(t-1)}|\mathbf{X}^{(t-)},\mathbf{A}^{(0)},y)$, respectively.}
\label{fig:framework}
\vspace{-1cm}
\end{figure}

% In order to take the advantages of both diffusion models and sub-graph GNNs, we finally choose the following decomposition format:
% choose to  

% By exploring the different decomposition, we find the following format enjoys both the advantages of generative diffusion models and sub-graph GNNs and detailed discussions can be 

% benefits of 

% easy fitting and good domain transfer capabilities.

\begin{equation}
    \label{eq:bayes}
    p(y|S) = p(y|\mathbf{A},\mathbf{X}) = \frac{p(\mathbf{X}|\mathbf{A},y)\cdot p(\mathbf{A}|y)\cdotp(y)}{\Sigma_{c\in \{0,1\}} p(\mathbf{A},\mathbf{X}|y=c) \cdot p(y=c)}
\end{equation}

\noindent where $p(y)$ is the prior distribution of node-pair's connection status over the graph. $p(\mathbf{A}|y)$ denotes the graph structure probability given the condition of nodes $u$ and $v$ being connected. $p(\mathbf{X}|\mathbf{A},y)$ represents the feature probability that is conditioned on the observed structure and connection status.  An overview of our framework is shown in Fig.~\ref{fig:framework}.  By splitting the generation of graph structure and node features into consequential steps, \method{} can be used for various link prediction settings, with or without node features. More importantly, since it integrates the idea of SGNNs, the structure diffusion model of \method{} can easily be transferred across datasets without involving any re-training procedure. In our experiments (Section~\ref{sec:transfer}), we find that small datasets can benefit from the learnt knowledge of larger datasets. Meanwhile, since \method{} has the feature component, we can design an independent feature diffusion model for each dataset to model diverse node features. In other words, the structure diffusion of \method{} provides a shareable basis for different datasets and the feature diffusion acts as an adjusting head which adapts the whole model to specific datasets. Lastly, generative models are well-known for their robustness against adversarial attacks ~\citep{wang2020deep,zimmermann2021score}. By modeling the likelihood scores of both the graph structure and node features, we expect that \method should be more robust against graph adversarial attacks as compared to existing discriminative approaches. We empirically verify this assumption  in Section~\ref{sec:robust-exp}. Next, we detail the major components of \method{}.

% In the following sections, we will introduce details about the likelihood score estimation on each component with diffusion models. And in Sec.~\ref{sec:advantages}, we will present the discussions about the advantages of our proposed method. \jt{we need to describe our overview framework. Also how this is related to the above description. Is the framework to illustrate how to calculate (1)?? I think we need more details and also connections between the formulation and figure}  The overview of our framework is shown as Fig.~\ref{fig:framework}.

% and detailed discussions are presented in Sec.~\ref{sec:advantages}

% With Eq.~\ref{eq:bayes}, the problem is decomposed and the estimations of each component are introduced in the following sections. 

% transformed to estimate the likelihood scores of each component.  

% calculate the likelihood score estimation to each component. 

% we transform the problem to calculate each component and fuse them details to estimate the categorical probability $\mathcal{P}(y_{uv})$. In the following sections, we will introduce the estimation of each component in details and the diagram of the framework is shown as Fig.~\ref{}.

\subsection{Structure Diffusion Model} \label{sec:graph-struct-diffusion}

The estimation of $p(\mathbf{A}|y)$ with diffusion models involves the discrete input $\mathbf{A}$. Following DiGress~\citep{vignac2022digress}, we use discrete status transition noise~\citep{austin2021structured} to maintain both the sparsity of the adjacency matrix as well as graph theoretic notions such as connectivity during the diffusion process.  In addition to the adjacency matrix $\mathbf{A}$, we further include the orbit features of each node $\mathbf{X}^{\prime}$ in the diffusion process. This is because we are estimating the likelihood score of a sub-graph $S$ under the connection condition $y$ of the sub-graph's center nodes $u$ and $v$. The orbit features indicate the relative distance of each node toward the center nodes thereby better distinguishing the sub-graphs with similar adjacency matrix but different center node locations. There are many choices for the orbit features. We use the Double Radius Node Labeling (DRNL)~\citep{zhang2021labeling} as we empirically find that it performs well on most of the datasets. To be concise, we define the forward of the structure diffusion model as:
\begin{align} 
    \label{eq:forward_process_digress}
    q(S^{(t)}|S^{(t-1)}) &= (\mathbf{A}^{(t-1)}\mathbf{Q}_A^t,\ \mathbf{X}^{{\prime}(t-1)}\mathbf{Q}_X^t),\nonumber\\ q(S^{(t)}|S^{(0)}) &= (\mathbf{A}^{(0)}\Bar{\mathbf{Q}}_A^t,\ \mathbf{X}^{{\prime}(0)}\Bar{\mathbf{Q}}_X^t),
\end{align}
where $\mathbf{Q}_A^t$ and $\mathbf{Q}_X^t$ are the transition probability matrices at the $t$-th step for discrete edge and node features. $\Bar{\mathbf{Q}}_A^t=\prod_{i=1}^t\mathbf{Q}_A^{i}$ and $\Bar{\mathbf{Q}}_X^t=\prod_{i=1}^t\mathbf{Q}_X^{i}$. The backward process can be stated as:
\begin{align}
\label{eq:backward}
    p_\theta(S^{(t-1)}|S^{(t)}, y) = (&\mathbf{A}^{(t)}(\mathbf{Q}^t_A)^{\prime} \odot \phi_\theta(\mathbf{A}^{(t)}, y,t)\Bar{\mathbf{Q}}^{t-1}_A,\nonumber\\
    &\mathbf{X}^{(t)}(\mathbf{Q}^t_X)^{\prime} \odot \phi_\theta(\mathbf{X}^{(t)}, y,t)\Bar{\mathbf{Q}}^{t-1}_X)
\end{align}
where $\odot$ denotes a element-wise product and $(\mathbf{Q}^t_A)^{\prime}$ and $(\mathbf{Q}^t_X)^{\prime}$ are the transpose of $\mathbf{Q}_A^t$ and $\mathbf{Q}_X^t$, respectively. $\phi_\theta$ is the denoising diffusion model, which takes timestep $t$, $t$-th step noisy sample $S^{(t)}$ and connection condition $y$ as inputs. It further outputs the distribution of categorical features in the clean graph $S^{(0)}$. We use a transformer-based neural network for $\phi_\theta$ and train it following prior work~\citep{vignac2022digress}. The conditional information $y$ is concatenated to every node and edge feature during the pre-processing step. 

The likelihood score of sub-graph $S$ can then be estimated by applying the evidence lower bound (ELBO) to the integration result of the joint probability $p_\theta(S^{(0:T)})=(\prod_{i=0}^T p_\theta(S^{(t-1)}|S^{(t)}))\cdot{P(S^{(T)})}$ over different trajectories $S^{(1:T)}$. We calculate $p_{\theta}(S|y)$ as follows:
\begin{align}
\label{eq:selbo}
    \log p_{\theta}(S|y) &\geq \log p(n_S|y) + \underbrace{D_{KL}[q(S^{(T)}|S)||q_{X}(n_S|y)\times q_{E}(n_S|y)]}_\textrm{Prior loss} \nonumber
    \\ 
    &+ \underbrace{\sum^T_{t=2}L_t(S|y)}_\textrm{Diffusion loss} + \underbrace{\mathbb{E}_{q(S^{(1)}|S)}[\log p_{\theta}(S|S^{(1)},y)]}_\textrm{Reconstruction loss}
\end{align}
with 
\begin{equation}
    L_t(S|y) = \mathbb{E}_{q(S^{(t)}|S)}[D_{KL}[q(S^{(t-1)}|S^{(t)},S)||p_{\theta}(S^{(t-1)}|S^{(t)},y)]]
\end{equation}
where $\log p(n_S|y)$ is the probability of sub-graph size $n_S$ under condition $y$. We note that since most link prediction problems are on undirected graphs, the above diffusion process is only applied to the upper-triangular of $\mathbf{A}$. Additionally, as the $(u,v)$-th element of adjacency matrix $\mathbf{A}^{(0)}$ will be unknown during test, we make $\mathbf{A}^{(0)}_{u,v}=0$ during pre-processing.

\subsection{Node Diffusion Model}
% \jt{can we use" Feature Diffusion model" to match figure 1??}

Since node features are typically continuous variables, we estimate $p_\theta(\mathbf{X}|\mathbf{A},y)$ using Gaussian noise as its diffusion kernel in a similar manner to DDPM~\citep{ho2020denoising}. Similar to Eq.~\eqref{eq:forward_process_digress}, the forward process of feature diffusion model can be written as:
\begin{align}
    q(\mathbf{X}^{(t)}|\mathbf{X}^{(t-1)}) &= \mathcal{N}(\mathbf{X}^{(t)};\sqrt{1-\beta_t}\mathbf{X}^{(t-1)},\beta_t\textbf{I}),\\
    q(\mathbf{X}^{(t)}|\mathbf{X}^{(0)}) &= \mathcal{N}(\mathbf{X}^{(t)};\sqrt{1-\Bar{\alpha}_t}\mathbf{X}^{(0)},\Bar{\alpha}_t\textbf{I}),
\end{align}
where $\beta_t$ is the variance schedule, which transitions from 0 to 1, and $\Bar{\alpha}_t=\prod_{i=1}^{t}(1-\beta_t)$. The reverse process under condition $c$ is defined as:
\begin{align}
    p_\theta(\mathbf{X}^{(t-1)}|\mathbf{X}^{(t)},c)= \mathcal{N}(&\mathbf{X}^{(t-1)};\Tilde{\mu}_t,\Tilde{\beta}_t)\nonumber\\
    \Tilde{\mu}_t=\frac{1}{\sqrt{\alpha_t}}(\mathbf{X}^{(t)}-\frac{1-\alpha_t}{\sqrt{1-\Bar{\alpha}_t}}\epsilon_\theta(\mathbf{X}^{(t)},t,c)&,\ \Tilde{\beta}_t=\frac{1-\Bar{\alpha}_{t-1}}{1-\Bar{\alpha}_t}\cdot\beta_t
\end{align}
where $\alpha_t=1-\beta_t$. $\epsilon_\theta(\mathbf{X}^{(t)},t,y)$ is our denoising diffusion models and $\Tilde{\beta}_t$ is only correlated with the $\beta_t$. Through applying the ELBO trick over the integral of joint distribution $q(\mathbf{X}^{(0:T)}|y)$, we can write the conditioned log-likelihood score of node features as:
\begin{align}
    &\log p_{\theta}(\mathbf{X}|c) \geq \mathbb{E}_q\left[\log\frac{p_\theta(\mathbf{X}^{(0:T)},c)}{q(\mathbf{X}^{(1:T)}|\mathbf{X}^{(0)})}\right]\nonumber \\ 
    &=\mathbb{E}_q[\underbrace{D_{KL}(q(\mathbf{X}^{(T)}|\mathbf{X}^{(0)})||p_\theta(\mathbf{X}^{(T)}))}_{\textrm{Prior loss}}  - \underbrace{\log p_\theta(\mathbf{X}^{(0)}|\mathbf{X}^{(1)},c)}_{\textrm{Reconstruction loss}} \nonumber\\
    &+ \underbrace{\Sigma^T_{t=2}D_{KL}(q(\mathbf{X}^{(t-1)}|\mathbf{X}^{(t)},\mathbf{X}^{(0)})||p_\theta(\mathbf{X}^{(t-1)}|\mathbf{X}^{(t)},c))}_{\textrm{Diffusion loss}}],
\end{align}
where the prior and reconstruction losses are nullified as their value is much smaller than the diffusion loss. To calculate $D_{KL}(q|p_\theta)$, we use the simplified form proposed by \citet{ho2020denoising} producing the final expression:
\begin{equation}
\label{eq:felbo}
    -\mathbb{E}_{t,\epsilon}[||\epsilon - \epsilon_{\theta}(\mathbf{X}^{(t)},c)||^2] \ \ \ \textrm{with}\ \ \mathbf{X}^{(t)} = \sqrt{\Bar{\alpha}_t}\mathbf{X}^{(0)} + \sqrt{1-\Bar{\alpha}_t}\epsilon ,
\end{equation}
where $\epsilon \sim \mathcal{N}(0,1)$. The denoising model $\epsilon_\theta$ takes as input $\epsilon_\theta$, the noisy samples at step $t$, and the given condition $c=({\bf A},y)$ and outputs the noise at step $t$. Since the condition $c$ includes both the adjacency matrix ${\bf A}$ and the connection condition $y$, the predictions are made at node-level. Lastly, we use GCN~\citep{kipf2016semi} to model $\epsilon_\theta$: 
\begin{equation}
    \epsilon_\theta(\mathbf{X}^{(t)},y,t) = \hat{{\bf A}}(\sigma(\hat{{\bf A}}{\bf X^{(t)}}{\bf W}_{0})){\bf W}_{1},
\end{equation}
where $\sigma$ is an activation function, and ${\bf W}_{0}$ and ${\bf W}_{1}$ are the learnable parameters. $\hat{{\bf A}}=\Tilde{{\bf D}}^{-\frac{1}{2}}\Tilde{{\bf A}}\Tilde{{\bf D}}^{-\frac{1}{2}}$ with $\Tilde{{\bf A}}={\bf A}+{\bf I}_N$ and $\Tilde{{\bf D}}_{ii}=\Sigma_j \Tilde{{\bf A}}_{ij}$. The connection status condition $y$ is concatenated to each node feature during the feature pre-processing. 

% Our model is described in details in Appendix~\ref{}. Finally, as Eq.~\ref{eq:felbo} is also used as the loss function for training the de-noising model, we directly use the calculated loss as the likelihood score estimations of our samples for the later prediction. 

% and reverse processes are Markov chain, the join probability of latent variables $x_{1:T}=\{x_1,x_2,...,x_t\}$ can be 

% And the 

% % \begin{equation}
    
% % \end{equation}

% which uses Gaussian noise as the diffusion kernel for its noise forward process. 

% Similar to the discrete diffusion model, the noise forward process of DDPMs is a Markov Chain and its 

% of our model can be written as Markov chain

% Different from the discrete diffusion noise kernel, applying Gaussian kernel

% applying Ga

% During the backward process, as our goal focuses on estimating the log-likelihood of given node features, under the condition of specific adjacency matrix ${\bf A}$ and the global connection label $y$, we introduce both ${\bf A}$ and $y$ as the additional inputs for node feature denoising neural network model $\phi_{\theta}$. The whole forward and backward process of our node feature diffusion model can be summarized as follows:

% designs as node features are usually continuous variables, e.g., semantic representation, 

% \hang{Introduce about how we estimate the node feature Log-Likelihood scores ($\mathcal{P}(X|A,Y)$).}

\subsection{Connection Probability Estimation}

With the estimated log-likelihood scores of the sample's graph structure $\log p({\bf A}|y)$ and node features $\log p(x|{\bf A},y)$, we can estimate the connection probability $P(y|{\bf A},{\bf X})$ via Eq.~\eqref{eq:bayes}. However, directly taking the summation over those two components will be sub-optimal, as the scale of values returned by the two diffusion models are different. Furthermore, the weighting values of the diffusion loss are neglected during the loss calculation for simplification purposes. Because of this, we use the additional learnable parameter set $\{\eta_1,\eta_2,\delta\}$ to flexibly adjust each component during the fusion. The final connection probability calculation can be written as:
\begin{equation}
    P(y|{\bf A},{\bf X}) =\textrm{softmax}_y(\log P({\bf A},{\bf X}|y) + \log P(y)),
\end{equation}
with 
\begin{equation}
    \log P({\bf A},{\bf X}|y) = \eta_1 \cdot \log \Hat{{\bf P}}({\bf X}|{\bf A},y) + \eta_2 \cdot \log \Hat{{\bf P}}({\bf A}|y) + \delta,
\end{equation}
where $\{\eta_1,\eta_2,\delta\}$ are optimized via gradient descent over the cross entropy loss between true links and the predicted connection probability $\Hat{P}(y|{\bf A}, {\bf X})$. Please check Appendix~\ref{app:pseudo_code} for more details.

%% file: experiment.tex
% \jt{In this section we aim to conduct comprehensive experiments to validate the advantages of the proposed framework \method in terms of xxx, xxx and xxx.  }
% In this section we aim to conduct comprehensive experiments to validate the advantages of the proposed framework \method in terms of cross-data transferability, robustness and generalization to limited-size training.

In this section we conduct comprehensive experiments to validate the advantages of the proposed framework \method. In particular, we aim to answer the following questions: \textbf{RQ1:} Does \method enjoy the advantages of both SGNNs and generative models in solving cross-dataset link prediction? \textbf{RQ2:} How is the generalization capability of \method when faced with the challenge of train size limitation? \textbf{RQ3:} Does \method show its strength in robustness against the adversarial attacks on graph structure? Before presenting our experimental results and observations, we first introduce our general experimental settings.

% \begin{compactitem}
%     \item \textbf{RQ1} Does \method enjoy the advantages of both SGNNs and generative models in solving cross-dataset link prediction? 
%     \item \textbf{RQ2} How is the generalization capability of \method when faced with the challenge of train size limitation?
%     \item \textbf{RQ3} Does \method show its strength in robustness against the adversarial attacks on graph structure? 
% \end{compactitem}

% \noindent Before presenting our experimental results and observations, we first introduce our general experimental settings.

\subsection{General Experimental Settings}

To demonstrate the effectiveness of \method, we choose seven representative link prediction algorithms as our baselines. Specifically, our baselines include GCN~\citep{kipf2016semi}, GAT~\citep{velivckovic2017graph}, SAGE~\citep{hamilton2017inductive}, NeoGNN~\citep{yun2021neo}, VGAE~\citep{kipf2016variational}, and SEAL~\cite{zhang2018link}. To be noticed, we select VGAE because it is a representative generative model for graph learning. And we collect SEAL and NeoGNNs since both of them are the effective link prediction models sharing the similar sub-graph learning ideas with \method. For the other baseline methods, we collect them following the prior studies on link prediction tasks~\citep{li2024evaluating}. More details about implementations of the baseline and \method can be found in Appendix~\ref{app:implement_detail}. We conduct experiments on six real-world graph datasets, including 3 citation networks: Cora, Citeseer and Pubmed~\cite{planetoid} and 3 miscellaneous networks: USAir, NS and Router. The details about each dataset are shown in Table~\ref{tab:data_stats}. Following prior works~\cite{kipf2016variational, zhang2018link}, we split the existing links in each graph into train/valid/test with the percentages 80\%/5\%/15\%. For evaluation, we randomly sample the same amount of unconnected node pairs as the negative samples. The evaluation metrics used in our experiment are AUC, Average Precision(AP) and Hit@100. All experiments are run over 10 seeds and we report both the mean values of each metric.

\begin{table}[!btph]
\centering
\caption{Performance on the cross-data link prediction tasks. The \textbf{Rank} displays the average rank of models in different source and target datasets. The best rank value is marked with $^{*}$, the second best is marked with $^{\ddagger}$, and the third best is marked with $^{\dagger}$.}
\label{tab:transfer}
\resizebox{\textwidth}{!}{
\begin{tabular}{@{}cccccccccccccc@{}}
\toprule
 &  & \multicolumn{6}{c}{Source} & \multicolumn{6}{c}{Target} \\ \cmidrule(l){3-14} 
\multirow{-2}{*}{Model} & \multirow{-2}{*}{Rank $\downarrow$} &  \ \ Cora \ \ & Citeseer & Pubmed & \ Router\ &\ \ \ NS\ \ \ & \ \ USAir \ \ & \ \ Cora\ \ & Citeseer & Pubmed &\ Router\ &\ \ \ NS\ \ \ &\ \ USAir\ \ \\ \midrule
\multicolumn{14}{c}{AUC $\uparrow$} \\ \midrule
\multicolumn{1}{c|}{GCN} & \multicolumn{1}{c|}{5.3} & \cellcolor[HTML]{EFEFEF}82.25 & 75.38 & \cellcolor[HTML]{EFEFEF}83.43 & 74.87 & \cellcolor[HTML]{EFEFEF}50.84 & 68.71 & \cellcolor[HTML]{EFEFEF}70.52 & 71.12 & \cellcolor[HTML]{EFEFEF}71.68 & 55.75 & \cellcolor[HTML]{EFEFEF}85.02 & 81.39 \\
\multicolumn{1}{c|}{GAT} & \multicolumn{1}{c|}{4.7} & \cellcolor[HTML]{EFEFEF}79.31 & 77.89 & \cellcolor[HTML]{EFEFEF}80.63 & 75.12 & \cellcolor[HTML]{EFEFEF}67.59 & 68.15 & \cellcolor[HTML]{EFEFEF}74.76 & 74.19 & \cellcolor[HTML]{EFEFEF}73.99 & 50.43 & \cellcolor[HTML]{EFEFEF}89.32 & 85.99 \\
\multicolumn{1}{c|}{SAGE} & \multicolumn{1}{c|}{4.0} & \cellcolor[HTML]{EFEFEF}82.41 & 81.28 & \cellcolor[HTML]{EFEFEF}84.17 & 76.30 & \cellcolor[HTML]{EFEFEF}70.63 & 68.14 & \cellcolor[HTML]{EFEFEF}73.33 & 73.01 & \cellcolor[HTML]{EFEFEF}79.54 & 65.45 & \cellcolor[HTML]{EFEFEF}85.81 & 85.78 \\
\multicolumn{1}{c|}{NeoGNN} & \multicolumn{1}{c|}{3.9$^{\dagger}$} & \cellcolor[HTML]{EFEFEF}80.26 & 74.49 & \cellcolor[HTML]{EFEFEF}85.99 & 81.09 & \cellcolor[HTML]{EFEFEF}63.83 & 78.21 & \cellcolor[HTML]{EFEFEF}82.67 & 78.05 & \cellcolor[HTML]{EFEFEF}82.48 & 53.37 & \cellcolor[HTML]{EFEFEF}90.15 & 77.16 \\
\multicolumn{1}{c|}{VGAE} & \multicolumn{1}{c|}{6.8} & \cellcolor[HTML]{EFEFEF}71.89 & 73.58 & \cellcolor[HTML]{EFEFEF}75.46 & 64.56 & \cellcolor[HTML]{EFEFEF}62.34 & 65.25 & \cellcolor[HTML]{EFEFEF}67.24 & 66.27 & \cellcolor[HTML]{EFEFEF}69.18 & 54.04 & \cellcolor[HTML]{EFEFEF}80.72 & 75.64 \\
\multicolumn{1}{c|}{SEAL} & \multicolumn{1}{c|}{1.9$^{\ddagger}$} & \cellcolor[HTML]{EFEFEF}89.09 & 84.55 & \cellcolor[HTML]{EFEFEF}88.84 & 87.98 & \cellcolor[HTML]{EFEFEF}86.55 & 75.03 & \cellcolor[HTML]{EFEFEF}84.88 & 83.14 & \cellcolor[HTML]{EFEFEF}80.02 & 78.45 & \cellcolor[HTML]{EFEFEF}94.86 & 90.69 \\
\multicolumn{1}{c|}{SGDiff} & \multicolumn{1}{c|}{1.4$^*$} & \cellcolor[HTML]{EFEFEF}85.94 & 90.49 & \cellcolor[HTML]{EFEFEF}92.07 & 87.99 & \cellcolor[HTML]{EFEFEF}87.98 & 83.80 & \cellcolor[HTML]{EFEFEF}86.93 & 86.23 & \cellcolor[HTML]{EFEFEF}90.78 & 88.62 & \cellcolor[HTML]{EFEFEF}91.62 & 84.09 \\ \midrule
\multicolumn{14}{c}{AP $\uparrow$} \\ \midrule
\multicolumn{1}{c|}{GCN} & \multicolumn{1}{c|}{5.5} & \cellcolor[HTML]{EFEFEF}83.75 & 79.37 & \cellcolor[HTML]{EFEFEF}85.94 & 74.88 & \cellcolor[HTML]{EFEFEF}58.78 & 69.11 & \cellcolor[HTML]{EFEFEF}73.67 & 73.00 & \cellcolor[HTML]{EFEFEF}74.37 & 62.85 & \cellcolor[HTML]{EFEFEF}85.83 & 82.11 \\
\multicolumn{1}{c|}{GAT} & \multicolumn{1}{c|}{4.4} & \cellcolor[HTML]{EFEFEF}81.44 & 80.68 & \cellcolor[HTML]{EFEFEF}83.57 & 80.10 & \cellcolor[HTML]{EFEFEF}72.31 & 72.21 & \cellcolor[HTML]{EFEFEF}78.27 & 79.40 & \cellcolor[HTML]{EFEFEF}76.66 & 58.39 & \cellcolor[HTML]{EFEFEF}92.88 & 84.72 \\
\multicolumn{1}{c|}{SAGE} & \multicolumn{1}{c|}{4.6} & \cellcolor[HTML]{EFEFEF}83.14 & 81.59 & \cellcolor[HTML]{EFEFEF}85.70 & 77.03 & \cellcolor[HTML]{EFEFEF}69.85 & 66.24 & \cellcolor[HTML]{EFEFEF}74.13 & 74.69 & \cellcolor[HTML]{EFEFEF}77.87 & 65.76 & \cellcolor[HTML]{EFEFEF}87.47 & 83.63 \\
\multicolumn{1}{c|}{NeoGNN} & \multicolumn{1}{c|}{3.3$^{\dagger}$} & \cellcolor[HTML]{EFEFEF}85.50 & 80.36 & \cellcolor[HTML]{EFEFEF}89.71 & 85.35 & \cellcolor[HTML]{EFEFEF}72.63 & 80.79 & \cellcolor[HTML]{EFEFEF}86.38 & 83.19 & \cellcolor[HTML]{EFEFEF}86.63 & 65.55 & \cellcolor[HTML]{EFEFEF}93.57 & 79.01 \\
\multicolumn{1}{c|}{VGAE} & \multicolumn{1}{c|}{6.9} & \cellcolor[HTML]{EFEFEF}71.21 & 74.21 & \cellcolor[HTML]{EFEFEF}76.21 & 64.81 & \cellcolor[HTML]{EFEFEF}61.52 & 65.07 & \cellcolor[HTML]{EFEFEF}67.56 & 66.01 & \cellcolor[HTML]{EFEFEF}69.38 & 57.23 & \cellcolor[HTML]{EFEFEF}78.35 & 74.49 \\
\multicolumn{1}{c|}{SEAL} & \multicolumn{1}{c|}{1.8$^{\ddagger}$} & \cellcolor[HTML]{EFEFEF}90.54 & 87.06 & \cellcolor[HTML]{EFEFEF}91.51 & 89.02 & \cellcolor[HTML]{EFEFEF}88.16 & 78.76 & \cellcolor[HTML]{EFEFEF}87.50 & 86.89 & \cellcolor[HTML]{EFEFEF}82.25 & 81.32 & \cellcolor[HTML]{EFEFEF}95.92 & 91.16 \\
\multicolumn{1}{c|}{SGDiff} & \multicolumn{1}{c|}{1.5$^*$} & \cellcolor[HTML]{EFEFEF}87.79 & 91.36 & \cellcolor[HTML]{EFEFEF}92.65 & 86.54 & \cellcolor[HTML]{EFEFEF}88.66 & 85.17 & \cellcolor[HTML]{EFEFEF}87.93 & 86.98 & \cellcolor[HTML]{EFEFEF}90.90 & 89.62 & \cellcolor[HTML]{EFEFEF}91.56 & 85.19 \\ \midrule
\multicolumn{14}{c}{Hit@100 $\uparrow$} \\ \midrule
\multicolumn{1}{c|}{GCN} & \multicolumn{1}{c|}{5.3} & \cellcolor[HTML]{EFEFEF}71.26 & 63.37 & \cellcolor[HTML]{EFEFEF}72.54 & 56.10 & \cellcolor[HTML]{EFEFEF}35.33 & 48.31 & \cellcolor[HTML]{EFEFEF}52.78 & 55.62 & \cellcolor[HTML]{EFEFEF}31.88 & 36.34 & \cellcolor[HTML]{EFEFEF}85.26 & 85.04 \\
\multicolumn{1}{c|}{GAT} & \multicolumn{1}{c|}{4.6} & \cellcolor[HTML]{EFEFEF}64.85 & 63.18 & \cellcolor[HTML]{EFEFEF}68.19 & 60.84 & \cellcolor[HTML]{EFEFEF}51.22 & 49.22 & \cellcolor[HTML]{EFEFEF}58.91 & 63.11 & \cellcolor[HTML]{EFEFEF}27.92 & 27.36 & \cellcolor[HTML]{EFEFEF}87.73 & 92.47 \\
\multicolumn{1}{c|}{SAGE} & \multicolumn{1}{c|}{4.8} & \cellcolor[HTML]{EFEFEF}68.06 & 65.35 & \cellcolor[HTML]{EFEFEF}73.38 & 57.45 & \cellcolor[HTML]{EFEFEF}48.49 & 43.62 & \cellcolor[HTML]{EFEFEF}54.19 & 58.28 & \cellcolor[HTML]{EFEFEF}26.50 & 39.79 & \cellcolor[HTML]{EFEFEF}85.16 & 92.44 \\
\multicolumn{1}{c|}{NeoGNN} & \multicolumn{1}{c|}{3.3$^{\dagger}$} & \cellcolor[HTML]{EFEFEF}72.09 & 65.13 & \cellcolor[HTML]{EFEFEF}77.77 & 70.12 & \cellcolor[HTML]{EFEFEF}52.23 & 63.56 & \cellcolor[HTML]{EFEFEF}74.96 & 69.35 & \cellcolor[HTML]{EFEFEF}48.35 & 41.76 & \cellcolor[HTML]{EFEFEF}89.78 & 76.70 \\
\multicolumn{1}{c|}{VGAE} & \multicolumn{1}{c|}{6.8} & \cellcolor[HTML]{EFEFEF}54.08 & 55.38 & \cellcolor[HTML]{EFEFEF}60.89 & 42.81 & \cellcolor[HTML]{EFEFEF}39.45 & 43.30 & \cellcolor[HTML]{EFEFEF}43.00 & 43.39 & \cellcolor[HTML]{EFEFEF}22.12 & 25.92 & \cellcolor[HTML]{EFEFEF}82.15 & 79.34 \\
\multicolumn{1}{c|}{SEAL} & \multicolumn{1}{c|}{1.8$^{\ddagger}$} & \cellcolor[HTML]{EFEFEF}78.42 & 74.46 & \cellcolor[HTML]{EFEFEF}82.31 & 77.71 & \cellcolor[HTML]{EFEFEF}73.46 & 59.26 & \cellcolor[HTML]{EFEFEF}77.16 & 76.14 & \cellcolor[HTML]{EFEFEF}39.09 & 62.92 & \cellcolor[HTML]{EFEFEF}94.96 & 95.35 \\
\multicolumn{1}{c|}{SGDiff} & \multicolumn{1}{c|}{1.5$^*$} & \cellcolor[HTML]{EFEFEF}74.99 & 83.37 & \cellcolor[HTML]{EFEFEF}86.02 & 76.82 & \cellcolor[HTML]{EFEFEF}75.01 & 74.51 & \cellcolor[HTML]{EFEFEF}78.95 & 80.08 & \cellcolor[HTML]{EFEFEF}50.43 & 78.67 & \cellcolor[HTML]{EFEFEF}93.84 & 88.75 \\ \bottomrule
\end{tabular}}
\vspace{-0.5cm}
\end{table}

\subsection{Performance on Cross-data Transferability}

\input{exp_transfer}

\begin{figure}[!btph]
    \centering
    \begin{subfigure}[b]{0.32\textwidth}
        \centering
        \includegraphics[width=\textwidth]{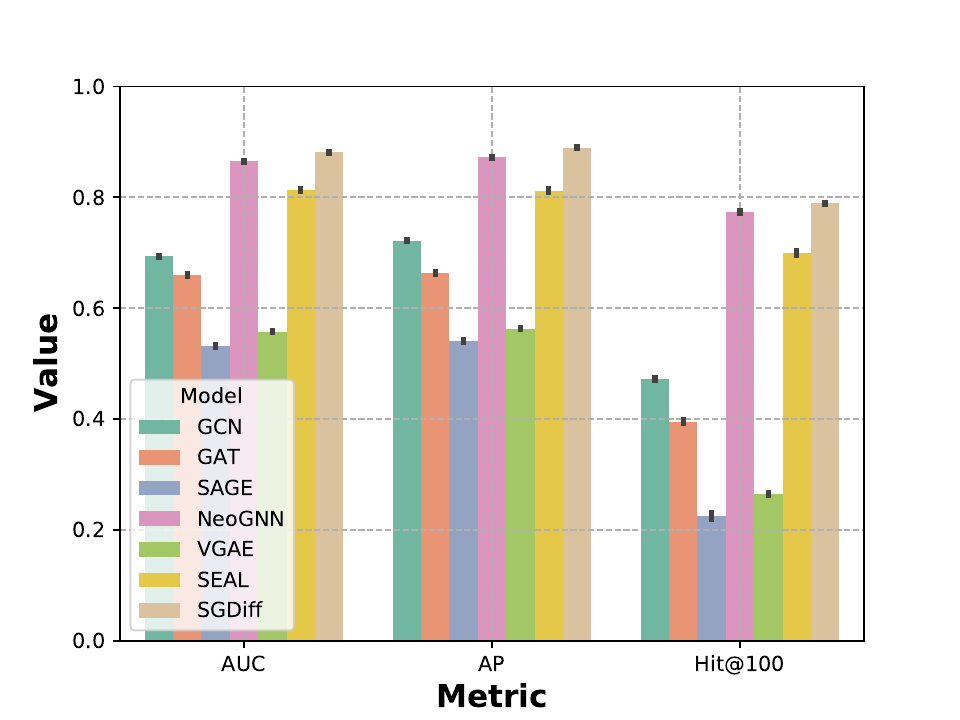}
        \caption{Cora}
        \label{fig:y equals x}
    \end{subfigure}
    \begin{subfigure}[b]{0.32\textwidth}
        \centering
        \includegraphics[width=\textwidth]{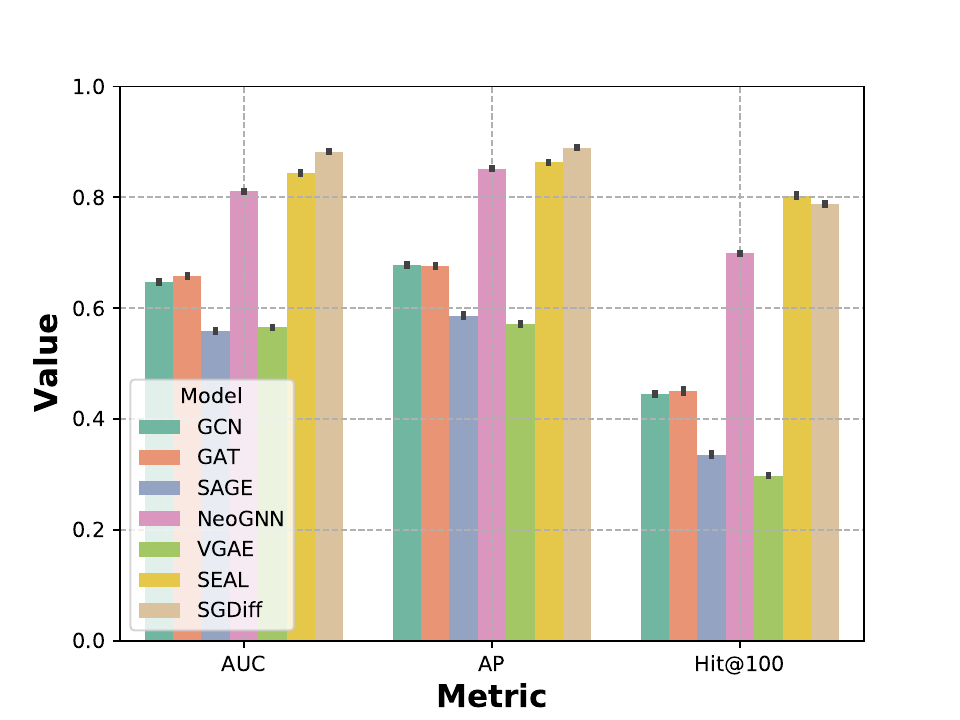}
        \caption{Citeseer}
        \label{fig:three sin x}
    \end{subfigure}
    \begin{subfigure}[b]{0.32\textwidth}
        \centering
        \includegraphics[width=\textwidth]{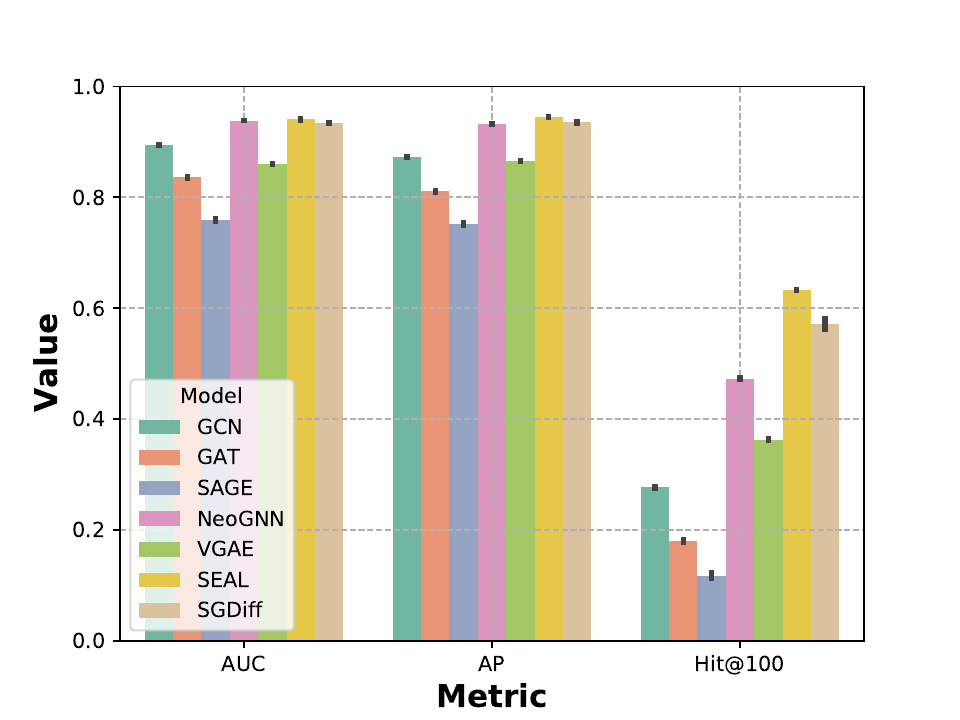}
        \caption{Pubmed}
        \label{fig:five over x}
    \end{subfigure}
     % \vspace{-0.5cm}
     \begin{subfigure}[b]{0.32\textwidth}
        \centering
        \includegraphics[width=\textwidth]{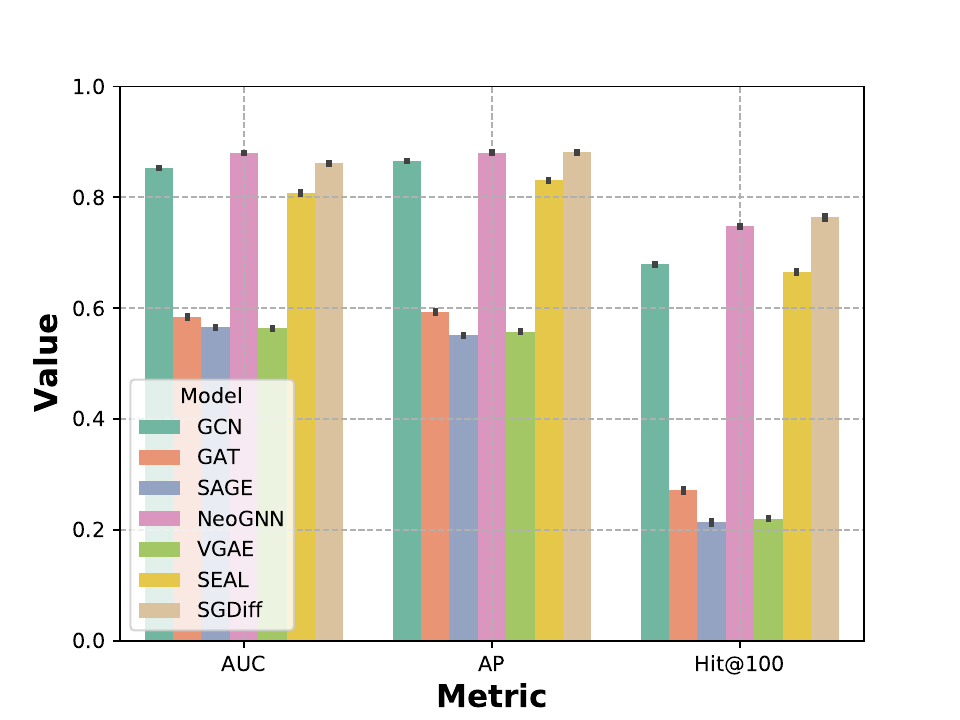}
        \caption{Router}
        \label{fig:y equals x}
    \end{subfigure}
    \begin{subfigure}[b]{0.32\textwidth}
        \centering
        \includegraphics[width=\textwidth]{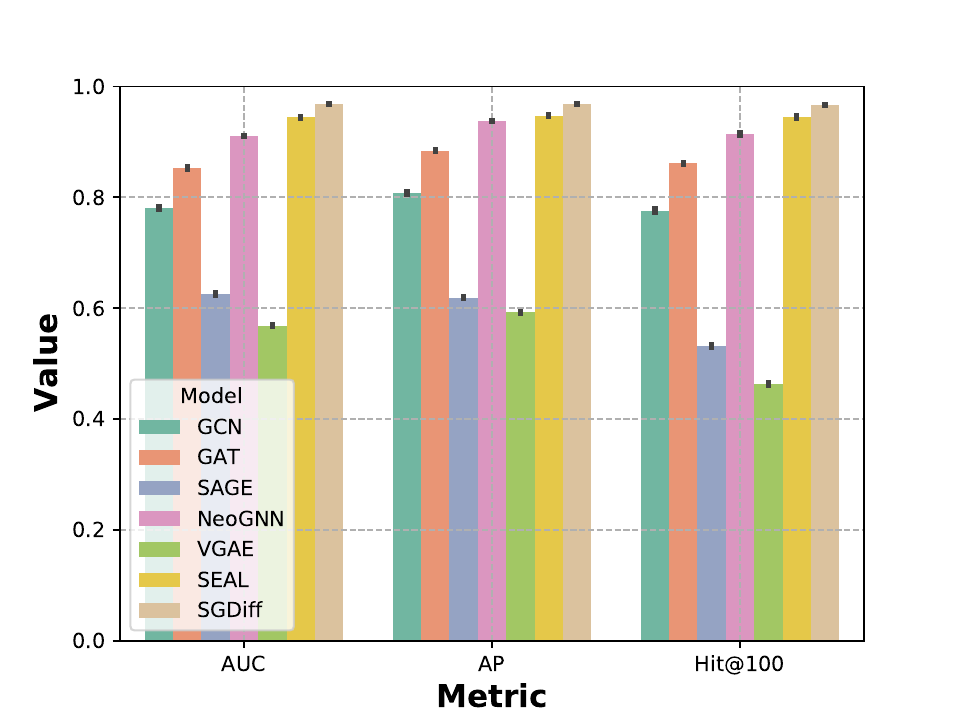}
        \caption{NS}
        \label{fig:three sin x}
    \end{subfigure}
    \begin{subfigure}[b]{0.32\textwidth}
        \centering
        \includegraphics[width=\textwidth]{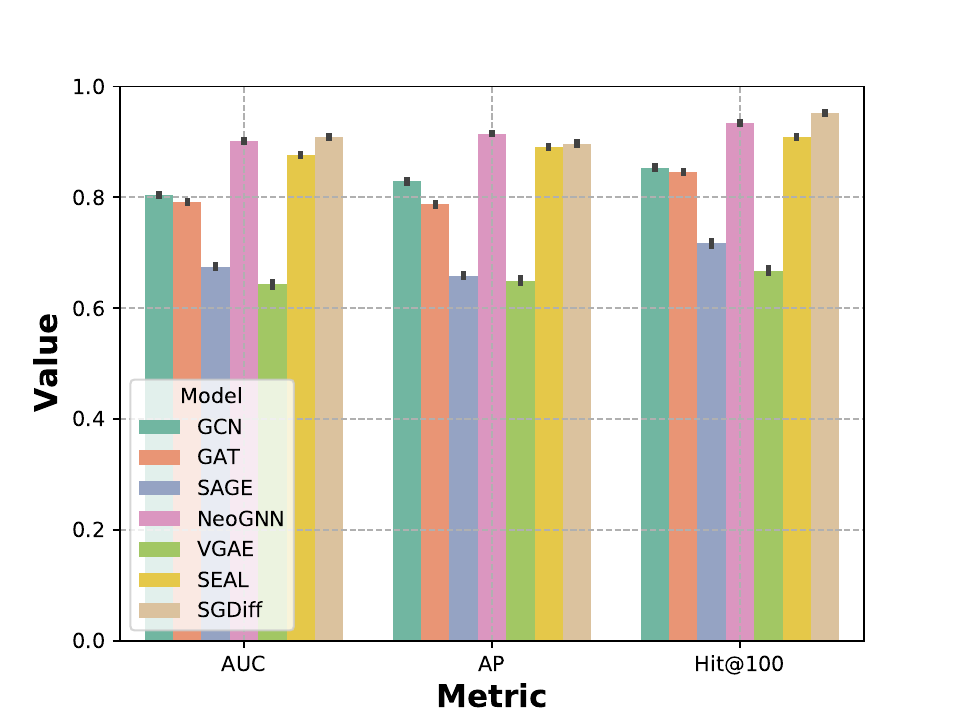}
        \caption{USAir}
        \label{fig:five over x}
    \end{subfigure}
    \caption{Model Performance on Cora / Citeseer / Pubmed / Router / NS / USAir datasets under the limited (1\%) training set scenario.}
    \label{fig:limit}
    \vspace{-0.5cm}
\end{figure}

\subsection{Performance with Train Size Constraint}
\label{sec:limit}

In this subsection, we further explore the generalization capability of \method and answer the second question by applying low availability limitations on the size of training set. In this setting, we shrink the training sample size of each dataset to only 1\%. To make the result comparable with the other experiments, only the size of the training data is decreased while keeping the validation and test sets. Furthermore, we use random sampling to create the smaller training sets. To be noticed, as we intend to explore the performance change caused by decreasing the training sample size, but not the completeness of the graph, we do not mask the remaining 99\% training edges from the original graph during the enclosing sub-graph generation process. We only control the number of sub-graphs used for training SEAL and \method. And for VGAE, we use the same adjacency matrix as other experiments but we mask 99\% of the cross-entropy loss over the adjacency matrix during the back propagation. The performance of \method and baseline models are shown in Figure~\ref{fig:limit}. We observe that as we limit the size of the training data, \method suffers less performance degradation compared with the other baseline models. This validates the strong capabilities of \method when little training data is used.

\subsection{Performance in Terms of Robustness}
\input{exp_robust}

%% file: exp_transfer.tex
\label{sec:transfer}
In this section, we aim to answer the first question about the cross-data transferablity of \method. As discussed in Section~\ref{sec:overview}, one potential advantage of the structure diffusion model of \method is the potential to be transferred across datasets without re-training. To validate this advantage, we perform a zero-shot cross dataset transferring experiment, where the model is trained with a source dataset and is tested on other target datasets. As the node features among different datasets are incompatible with each other, we do not add node features for \method and SEAL. For VGAE, as the training and test graphs have different node numbers, we do not use node-id as input features for VGAE. Instead, we follow prior work~\cite{wang2022equivariant}, which randomly projects the node features into the same dimension and then performs row normalization. For graphs without node features, we draw random vectors from the Gaussian distribution and use it as node features. We test the transferability by setting each of the six graph datasets as the source for training and test the train model overall all six graphs. We report the perforamnce of each model from two perspectives, \textbf{Source} and \textbf{Target}. To be specific, \textbf{Source} averages the test performance on different target datasets of one model trained with one fixed source dataset. And \textbf{Target} averages the test performances on one fixed target dataset of six  models trained by different source datasets. Overall, the cross-data transferring results are shown as Table~\ref{tab:transfer}, and we calculate the average rank of each model under different source and target dataset as the indicator for model's cross-data transferability. Detailed performance of models can be found in Appendix.~\ref{app:crossperform_detail}.

From Table~\ref{tab:transfer} we have the following observations: (1) the link prediction performance of all baseline models is always much better than the random guess, e.g., AUC greater than 50\%. This fact indicates that different graph datasets actually share some similar structure patterns for link prediction task and it will be possible to develop a unified link prediction model across different graph datasets. (2) Compared to the other baseline models, VGAE always receives the worst performance in cross-data transferring test. This phenomenon is consistent with our prior analysis on the poor transferring defects of current generative graph learning methods, which caused by neglecting the node feature reconstruction design. (3) \method achieves best performance in most of the transferring scenarios, which supports our claim that \method takes the advantages of both SGNNs and generative model in generalization. Additionally, we find that \method consistently benefits from transferring from a larger source dataset. For instance, when trained with different source datasets, \method receives the best performance with Pubmed, which indicates that fitting on Pubmed tends to produce the best transferability. In addition, this observation is followed by the other five source datasets, Cora, Router, Citeseer, NS and USAir, where USAir is the smallest. This phenomenon encourages us to explore an unified pre-training framework for link prediction as one promising future direction. 

% most datasets' metrics receive improvements while using Pubmed as source dataset. 

% Additionally, \method consistently benefits from transferring from a larger source dataset, e.g., most datasets' metrics receive improvements while using Pubmed as source dataset, this demonstrate the great potential \method in leverage larger graph dataset and encourage us to explore an unified pre-training framework for link prediction in the future.

% and has the great potential to be developed as the unified link prediction model for different graph dataset. (4) By views the 

% also consistent will our 

% can observe that \method achieves strong performance in most of the transferring scenarios, which support our claim that \method inherit the advantages of generative framework and SGNNs in generalization. 

% Additionally, we find that \method consistently benefits from transferring from a larger source dataset, see most datasets' metrics receive improvements while using Pubmed as source dataset. Fitting on Pubmed tends to produce the best transferability. This is followed by the other two datasets Citeseer and Cora where Cora is the smallest. This phenomenon encourages us to explore an unified pre-training framework for link prediction in the future. 

%% file: exp_robust.tex
\label{sec:robust-exp}

\begin{table}[]
\centering
\caption{Robustness against random flip (RF) and node embedding (NE) Attacks on Cora / Citeseer / Pubmed. The \textbf{Rank} displays the average rank of models in different settings. The best rank value is marked with $^{*}$, the second best is marked with $^{\ddagger}$, and the third best is marked with $^{\dagger}$.}
\label{tab:robust}
\resizebox{\textwidth}{!}{
\begin{tabular}{@{}cccccccccccccc@{}}
\toprule
 &  & \multicolumn{4}{c}{Cora} & \multicolumn{4}{c}{Citeseer} & \multicolumn{4}{c}{Pubmed} \\ \cmidrule(l){3-14} 
 &  & \multicolumn{2}{c}{RF} & \multicolumn{2}{c}{EA} & \multicolumn{2}{c}{RF} & \multicolumn{2}{c}{EA} & \multicolumn{2}{c}{RF} & \multicolumn{2}{c}{EA} \\ \cmidrule(l){3-14} 
\multirow{-3}{*}{Model} & \multirow{-3}{*}{Rank $\downarrow$} & \multicolumn{1}{c}{\ \ \ 25\%\ \ \ } & \multicolumn{1}{c}{\ \ \ 50\%\ \ \ } & \multicolumn{1}{c}{\ \ \ 25\%\ \ \ } & \multicolumn{1}{c}{\ \ \ 50\%\ \ \ } & \multicolumn{1}{c}{\ \ \ 25\%\ \ \ } & \multicolumn{1}{c}{\ \ \ 50\%\ \ \ } & \multicolumn{1}{c}{\ \ \ 25\%\ \ \ } & \multicolumn{1}{c}{\ \ \ 50\%\ \ \ } & \multicolumn{1}{c}{\ \ \ 25\%\ \ \ } & \multicolumn{1}{c}{\ \ \ 50\%\ \ \ } & \multicolumn{1}{c}{\ \ \ 25\%\ \ \ } & \multicolumn{1}{c}{\ \ \ 50\%\ \ \ } \\ \midrule
\multicolumn{14}{c}{AUC $\uparrow$} \\ \midrule
\multicolumn{1}{c|}{GCN} & \multicolumn{1}{c|}{5.4} & \cellcolor[HTML]{EFEFEF}84.66 & 82.29 & \cellcolor[HTML]{EFEFEF}84.23 & 84.23 & \cellcolor[HTML]{EFEFEF}80.55 & 78.63 & \cellcolor[HTML]{EFEFEF}80.93 & 79.60 & \cellcolor[HTML]{EFEFEF}95.04 & 93.28 & \cellcolor[HTML]{EFEFEF}94.10 & 93.10 \\
\multicolumn{1}{c|}{GAT} & \multicolumn{1}{c|}{5.2} & \cellcolor[HTML]{EFEFEF}86.90 & 83.60 & \cellcolor[HTML]{EFEFEF}87.53 & 84.23 & \cellcolor[HTML]{EFEFEF}84.19 & 81.31 & \cellcolor[HTML]{EFEFEF}84.90 & 82.64 & \cellcolor[HTML]{EFEFEF}88.92 & 84.72 & \cellcolor[HTML]{EFEFEF}88.71 & 83.86 \\
\multicolumn{1}{c|}{SAGE} & \multicolumn{1}{c|}{2.7$^{\ddagger}$} & \cellcolor[HTML]{EFEFEF}89.63 & 86.75 & \cellcolor[HTML]{EFEFEF}87.97 & 85.54 & \cellcolor[HTML]{EFEFEF}86.16 & 84.53 & \cellcolor[HTML]{EFEFEF}86.87 & 85.89 & \cellcolor[HTML]{EFEFEF}94.88 & 92.43 & \cellcolor[HTML]{EFEFEF}94.54 & 91.56 \\
\multicolumn{1}{c|}{NeoGNN} & \multicolumn{1}{c|}{3.7} & \cellcolor[HTML]{EFEFEF}89.15 & 86.59 & \cellcolor[HTML]{EFEFEF}87.51 & 84.92 & \cellcolor[HTML]{EFEFEF}82.59 & 80.96 & \cellcolor[HTML]{EFEFEF}83.00 & 82.19 & \cellcolor[HTML]{EFEFEF}94.40 & 93.68 & \cellcolor[HTML]{EFEFEF}95.30 & 93.21 \\
\multicolumn{1}{c|}{VGAE} & \multicolumn{1}{c|}{3.0$^{\dagger}$} & \cellcolor[HTML]{EFEFEF}88.87 & 86.61 & \cellcolor[HTML]{EFEFEF}87.38 & 85.05 & \cellcolor[HTML]{EFEFEF}90.07 & 87.46 & \cellcolor[HTML]{EFEFEF}89.96 & 87.45 & \cellcolor[HTML]{EFEFEF}94.35 & 92.42 & \cellcolor[HTML]{EFEFEF}94.24 & 92.27 \\
\multicolumn{1}{c|}{SEAL} & \multicolumn{1}{c|}{6.2} & \cellcolor[HTML]{EFEFEF}86.84 & 83.49 & \cellcolor[HTML]{EFEFEF}87.03 & 82.84 & \cellcolor[HTML]{EFEFEF}82.66 & 78.55 & \cellcolor[HTML]{EFEFEF}82.70 & 79.26 & \cellcolor[HTML]{EFEFEF}91.90 & 87.39 & \cellcolor[HTML]{EFEFEF}90.71 & 86.07 \\
\multicolumn{1}{c|}{SGDiff} & \multicolumn{1}{c|}{1.8$^{*}$} & \cellcolor[HTML]{EFEFEF}88.37 & 86.88 & \cellcolor[HTML]{EFEFEF}88.00 & 85.67 & \cellcolor[HTML]{EFEFEF}87.13 & 85.43 & \cellcolor[HTML]{EFEFEF}86.51 & 84.82 & \cellcolor[HTML]{EFEFEF}95.09 & 94.62 & \cellcolor[HTML]{EFEFEF}94.76 & 93.99 \\ \midrule
\multicolumn{14}{c}{AP $\uparrow$} \\ \midrule
\multicolumn{1}{c|}{GCN} & \multicolumn{1}{c|}{5.6} & \cellcolor[HTML]{EFEFEF}85.79 & 83.37 & \cellcolor[HTML]{EFEFEF}85.11 & 85.11 & \cellcolor[HTML]{EFEFEF}81.33 & 80.06 & \cellcolor[HTML]{EFEFEF}81.74 & 81.36 & \cellcolor[HTML]{EFEFEF}95.18 & 93.51 & \cellcolor[HTML]{EFEFEF}95.25 & 93.47 \\
\multicolumn{1}{c|}{GAT} & \multicolumn{1}{c|}{6.1} & \cellcolor[HTML]{EFEFEF}87.33 & 84.00 & \cellcolor[HTML]{EFEFEF}88.41 & 84.79 & \cellcolor[HTML]{EFEFEF}85.90 & 83.32 & \cellcolor[HTML]{EFEFEF}86.68 & 84.78 & \cellcolor[HTML]{EFEFEF}89.28 & 85.28 & \cellcolor[HTML]{EFEFEF}89.15 & 84.40 \\
\multicolumn{1}{c|}{SAGE} & \multicolumn{1}{c|}{3.1$^{\dagger}$} & \cellcolor[HTML]{EFEFEF}89.79 & 87.98 & \cellcolor[HTML]{EFEFEF}90.52 & 89.08 & \cellcolor[HTML]{EFEFEF}87.25 & 85.77 & \cellcolor[HTML]{EFEFEF}87.86 & 86.99 & \cellcolor[HTML]{EFEFEF}95.27 & 93.08 & \cellcolor[HTML]{EFEFEF}95.04 & 92.51 \\
\multicolumn{1}{c|}{NeoGNN} & \multicolumn{1}{c|}{2.5$^{*}$} & \cellcolor[HTML]{EFEFEF}90.84 & 89.25 & \cellcolor[HTML]{EFEFEF}91.05 & 89.82 & \cellcolor[HTML]{EFEFEF}85.77 & 84.53 & \cellcolor[HTML]{EFEFEF}86.16 & 85.36 & \cellcolor[HTML]{EFEFEF}95.59 & 94.06 & \cellcolor[HTML]{EFEFEF}95.54 & 93.83 \\
\multicolumn{1}{c|}{VGAE} & \multicolumn{1}{c|}{2.7$^{\ddagger}$} & \cellcolor[HTML]{EFEFEF}90.12 & 88.18 & \cellcolor[HTML]{EFEFEF}88.75 & 86.72 & \cellcolor[HTML]{EFEFEF}91.09 & 88.95 & \cellcolor[HTML]{EFEFEF}91.21 & 89.07 & \cellcolor[HTML]{EFEFEF}94.94 & 93.52 & \cellcolor[HTML]{EFEFEF}94.90 & 93.44 \\
\multicolumn{1}{c|}{SEAL} & \multicolumn{1}{c|}{4.9} & \cellcolor[HTML]{EFEFEF}89.34 & 86.89 & \cellcolor[HTML]{EFEFEF}89.72 & 86.65 & \cellcolor[HTML]{EFEFEF}86.83 & 83.73 & \cellcolor[HTML]{EFEFEF}86.96 & 84.56 & \cellcolor[HTML]{EFEFEF}93.15 & 89.88 & \cellcolor[HTML]{EFEFEF}92.40 & 89.28 \\
\multicolumn{1}{c|}{SGDiff} & \multicolumn{1}{c|}{3.2} & \cellcolor[HTML]{EFEFEF}88.35 & 87.10 & \cellcolor[HTML]{EFEFEF}88.09 & 86.04 & \cellcolor[HTML]{EFEFEF}89.17 & 87.79 & \cellcolor[HTML]{EFEFEF}88.65 & 87.46 & \cellcolor[HTML]{EFEFEF}95.12 & 94.62 & \cellcolor[HTML]{EFEFEF}94.97 & 94.47 \\ \midrule
\multicolumn{14}{c}{Hit@100 $\uparrow$} \\ \midrule
\multicolumn{1}{c|}{GCN} & \multicolumn{1}{c|}{6.2} & \cellcolor[HTML]{EFEFEF}72.73 & 67.40 & \cellcolor[HTML]{EFEFEF}71.88 & 71.88 & \cellcolor[HTML]{EFEFEF}68.97 & 64.16 & \cellcolor[HTML]{EFEFEF}68.78 & 66.21 & \cellcolor[HTML]{EFEFEF}64.06 & 56.44 & \cellcolor[HTML]{EFEFEF}64.13 & 57.15 \\
\multicolumn{1}{c|}{GAT} & \multicolumn{1}{c|}{5.8} & \cellcolor[HTML]{EFEFEF}77.48 & 70.34 & \cellcolor[HTML]{EFEFEF}78.52 & 71.51 & \cellcolor[HTML]{EFEFEF}73.73 & 69.83 & \cellcolor[HTML]{EFEFEF}74.83 & 71.01 & \cellcolor[HTML]{EFEFEF}43.32 & 33.70 & \cellcolor[HTML]{EFEFEF}42.72 & 32.15 \\
\multicolumn{1}{c|}{SAGE} & \multicolumn{1}{c|}{2.3$^{\ddagger}$} & \cellcolor[HTML]{EFEFEF}82.28 & 78.18 & \cellcolor[HTML]{EFEFEF}82.37 & 80.25 & \cellcolor[HTML]{EFEFEF}77.08 & 73.45 & \cellcolor[HTML]{EFEFEF}77.67 & 75.61 & \cellcolor[HTML]{EFEFEF}67.65 & 57.80 & \cellcolor[HTML]{EFEFEF}66.67 & 56.47 \\
\multicolumn{1}{c|}{NeoGNN} & \multicolumn{1}{c|}{3.9} & \cellcolor[HTML]{EFEFEF}80.44 & 77.37 & \cellcolor[HTML]{EFEFEF}81.04 & 78.27 & \cellcolor[HTML]{EFEFEF}70.00 & 69.08 & \cellcolor[HTML]{EFEFEF}70.52 & 69.97 & \cellcolor[HTML]{EFEFEF}63.98 & 58.61 & \cellcolor[HTML]{EFEFEF}63.83 & 59.45 \\
\multicolumn{1}{c|}{VGAE} & \multicolumn{1}{c|}{2.2$^{*}$} & \cellcolor[HTML]{EFEFEF}80.49 & 76.94 & \cellcolor[HTML]{EFEFEF}78.59 & 73.87 & \cellcolor[HTML]{EFEFEF}84.30 & 79.50 & \cellcolor[HTML]{EFEFEF}82.92 & 79.13 & \cellcolor[HTML]{EFEFEF}67.58 & 62.81 & \cellcolor[HTML]{EFEFEF}68.23 & 62.86 \\
\multicolumn{1}{c|}{SEAL} & \multicolumn{1}{c|}{5.1} & \cellcolor[HTML]{EFEFEF}78.58 & 72.65 & \cellcolor[HTML]{EFEFEF}78.80 & 72.11 & \cellcolor[HTML]{EFEFEF}74.98 & 68.22 & \cellcolor[HTML]{EFEFEF}74.43 & 68.78 & \cellcolor[HTML]{EFEFEF}63.87 & 56.65 & \cellcolor[HTML]{EFEFEF}63.73 & 58.49 \\
\multicolumn{1}{c|}{SGDiff} & \multicolumn{1}{c|}{2.5$^{\dagger}$} & \cellcolor[HTML]{EFEFEF}79.55 & 77.10 & \cellcolor[HTML]{EFEFEF}78.82 & 74.53 & \cellcolor[HTML]{EFEFEF}79.49 & 76.08 & \cellcolor[HTML]{EFEFEF}78.11 & 74.48 & \cellcolor[HTML]{EFEFEF}65.63 & 63.27 & \cellcolor[HTML]{EFEFEF}65.51 & 64.42 \\ \bottomrule
\end{tabular}}
\vspace{-0.5cm}
\end{table}

\begin{figure}[!btph]
     \centering
     \begin{subfigure}[b]{0.3\textwidth}
         \centering
         \includegraphics[width=\textwidth]{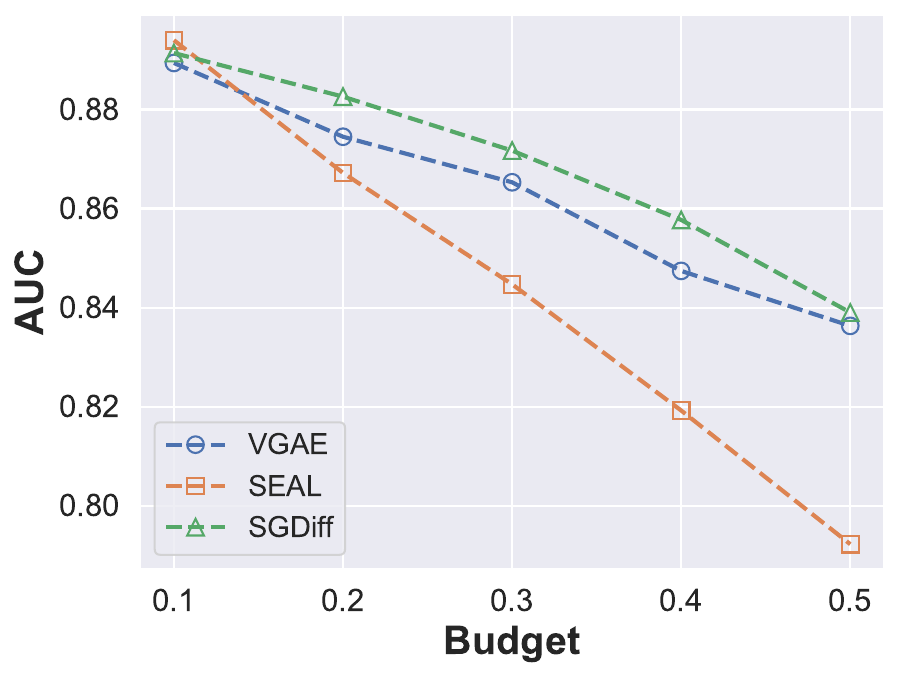}
         \caption{AUC on Cora}
         \label{fig:y equals x}
     \end{subfigure}
     \hspace{0.1cm}
     \begin{subfigure}[b]{0.3\textwidth}
         \centering
         \includegraphics[width=\textwidth]{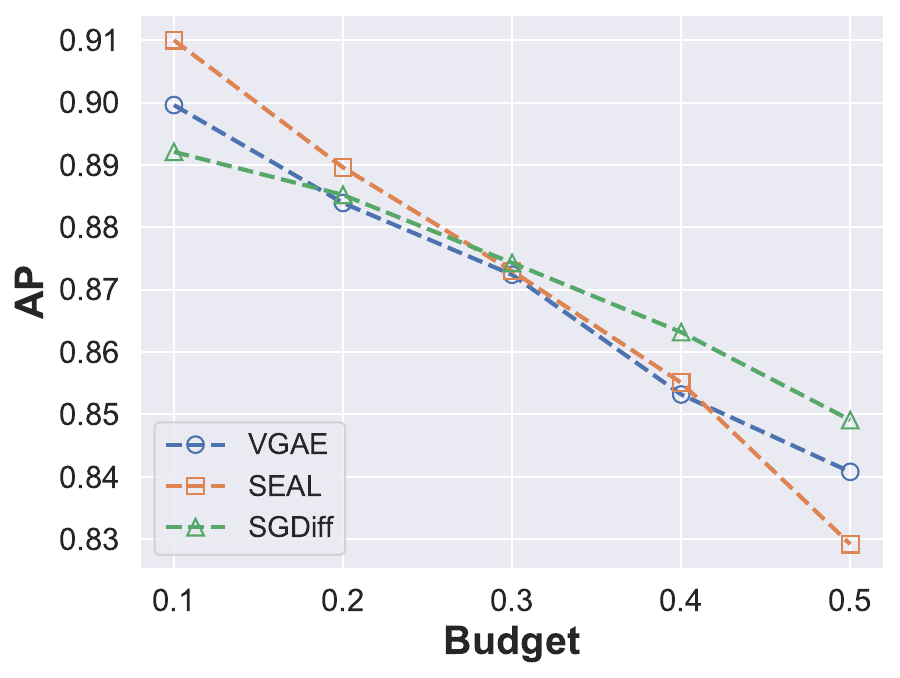}
         \caption{AP on Cora}
         \label{fig:three sin x}
     \end{subfigure}
     \hspace{0.1cm}
     \begin{subfigure}[b]{0.3\textwidth}
         \centering
         \includegraphics[width=\textwidth]{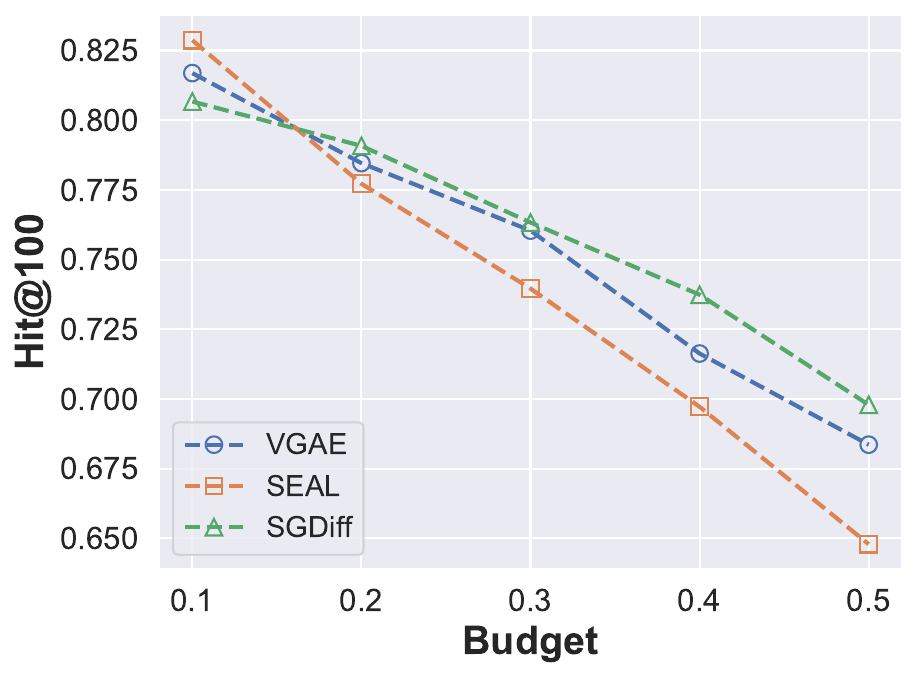}
         \caption{Hit@100 on Cora}
         \label{fig:five over x}
     \end{subfigure}
     \begin{subfigure}[b]{0.3\textwidth}
         \centering
         \includegraphics[width=\textwidth]{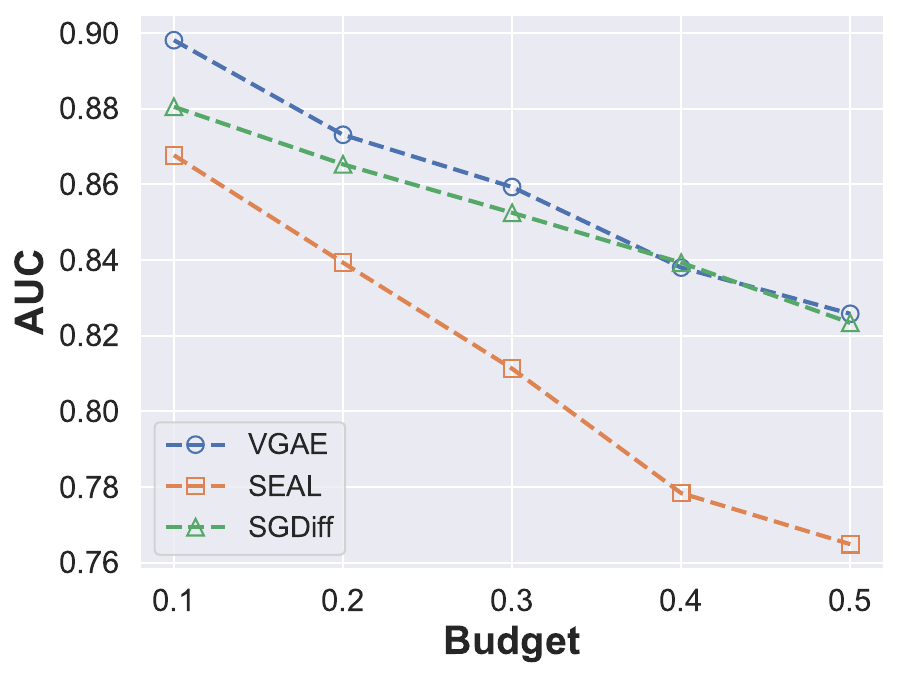}
         \caption{AUC on Citeseer}
         \label{fig:y equals x}
     \end{subfigure}
     \hspace{0.1cm}
     \begin{subfigure}[b]{0.3\textwidth}
         \centering
         \includegraphics[width=\textwidth]{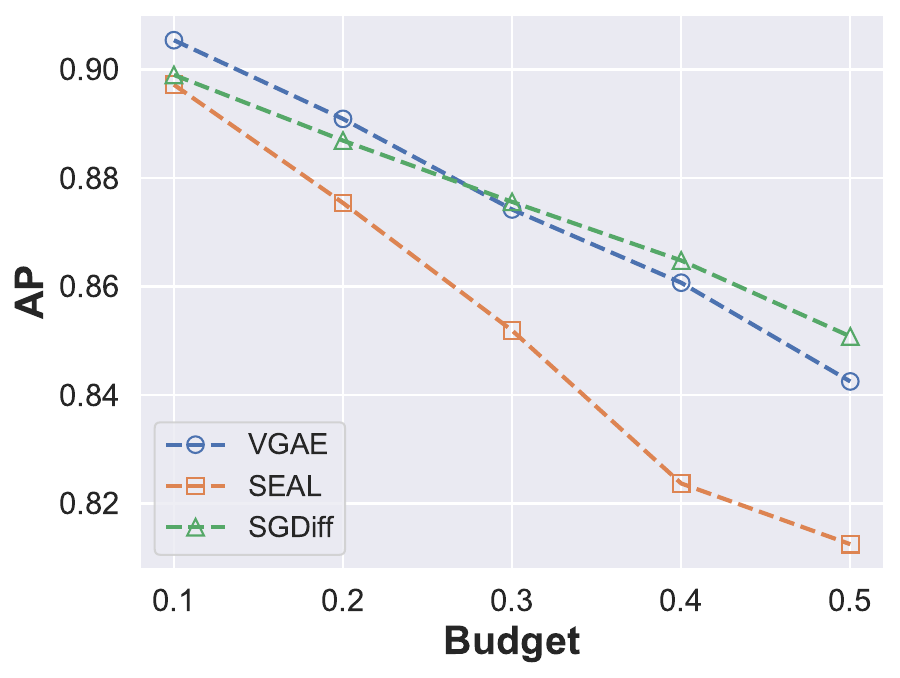}
         \caption{AP on Citeseer}
         \label{fig:three sin x}
     \end{subfigure}
     \hspace{0.1cm}
     \begin{subfigure}[b]{0.3\textwidth}
         \centering
         \includegraphics[width=\textwidth]{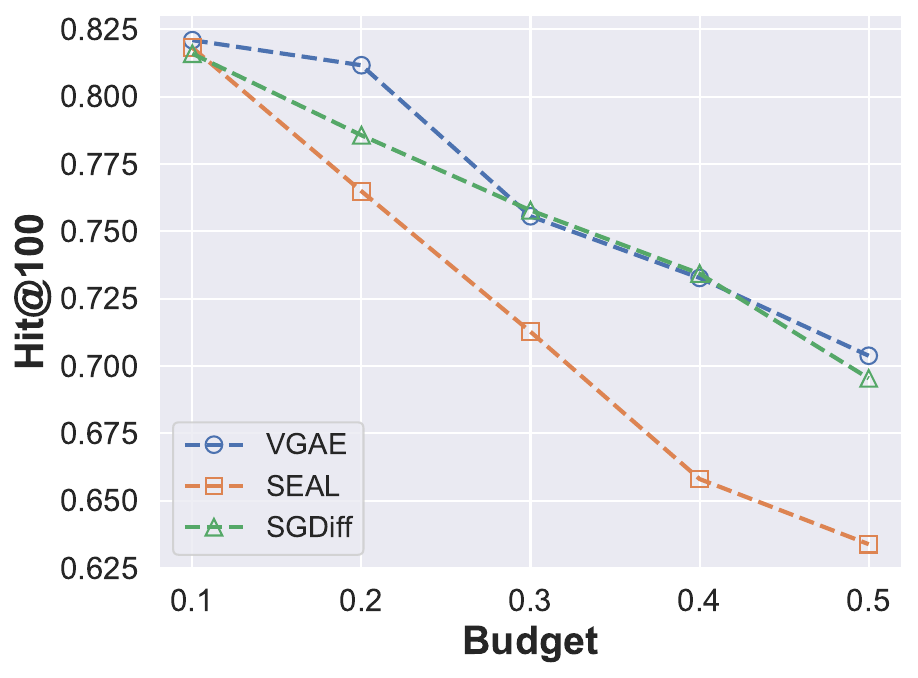}
         \caption{Hit@100 on Citeseer}
         \label{fig:five over x}
     \end{subfigure}
     \begin{subfigure}[b]{0.3\textwidth}
         \centering
         \includegraphics[width=\textwidth]{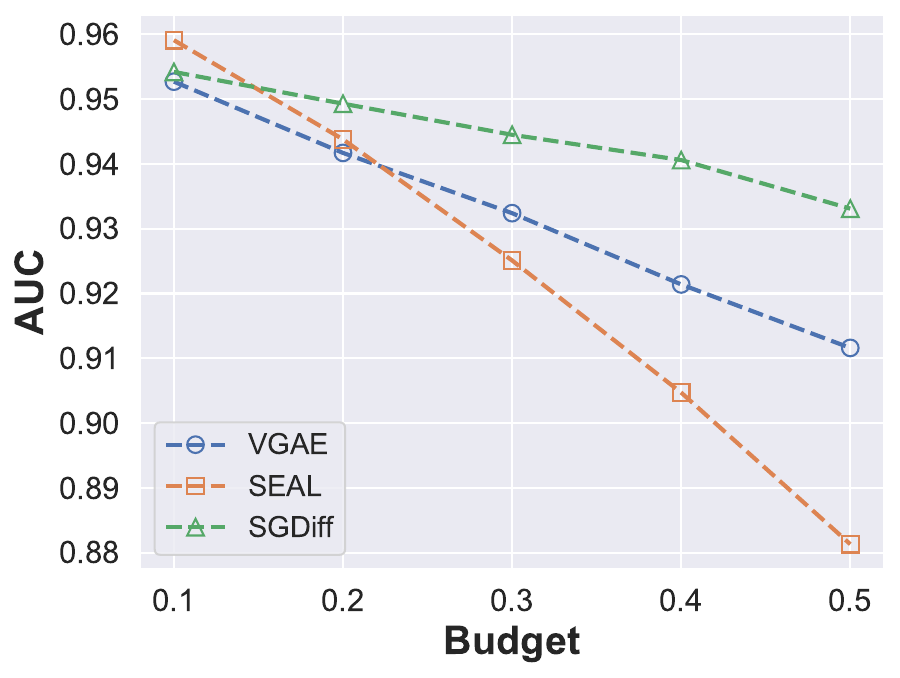}
         \caption{AUC on Pubmed}
         \label{fig:y equals x}
     \end{subfigure}
     \hspace{0.1cm}
     \begin{subfigure}[b]{0.3\textwidth}
         \centering
         \includegraphics[width=\textwidth]{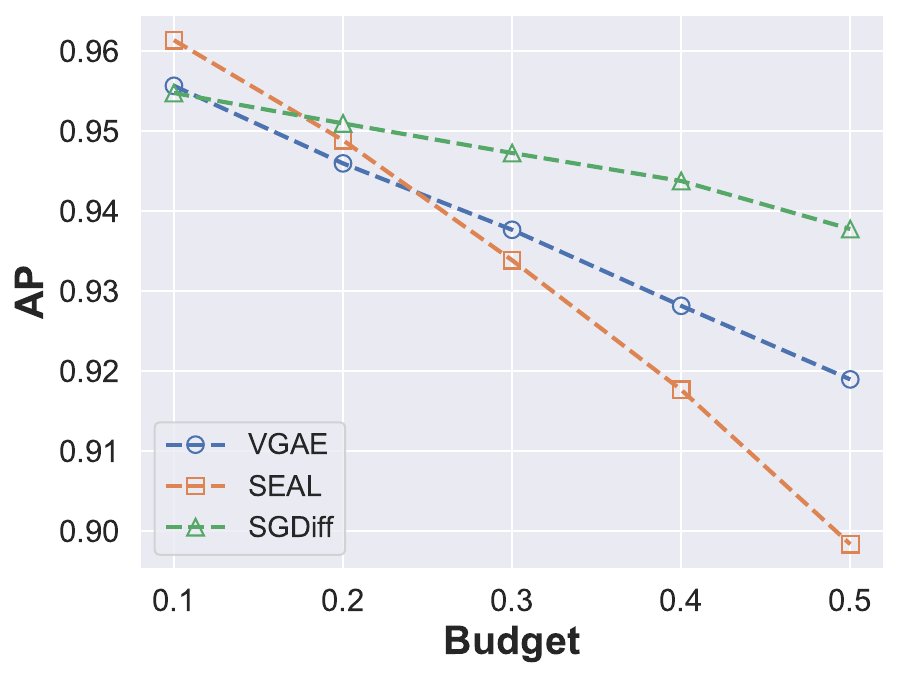}
         \caption{AP on Pubmed}
         \label{fig:three sin x}
     \end{subfigure}
     \hspace{0.1cm}
     \begin{subfigure}[b]{0.3\textwidth}
         \centering
         \includegraphics[width=\textwidth]{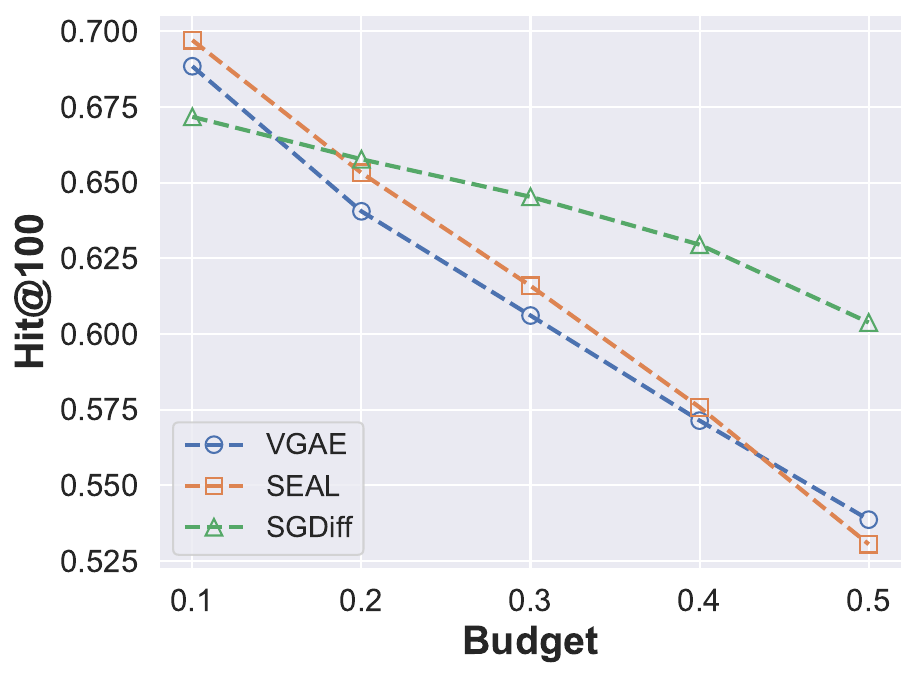}
         \caption{Hit@100 on Pubmed}
         \label{fig:five over x}
     \end{subfigure}
        \caption{Models' robustness against the DICE attack on Cora / Citeseer / Pubmed datasets.}
        \label{fig:robust}
        \vspace{-0.5cm}
\end{figure}

% RF and EA do not leverage node label information to generate attacked graph, we apply them with all six datasets.

In this section, we answer the third question by demonstrating the robustness of \method. To empirically test this, we adopt three common adversarial attack baselines for link predictions, i.e., random flipping (RF), Embedding Attack (EA)~\cite{bojchevski2019adversarial} and DICE~\cite{zhang2022unsupervised}. To be noticed, as most adversarial attack are proposed for graphs with node features, we conduct the following experiments with three citation networks: Cora, Citeseer and Pubmed. For DICE, as it required node label information as the supervised signal to train a surrogate model during the attack process, we apply it only with two baseline models, e.g., VGAE, SEAL, and \method. The implementation of each attack uses the open source graph attack tool library, DeepRobust~\cite{li2020deeprobust}. For each type of attacks, we substitute the clean adjacency matrix with an attacked one during the inference process. We then compare each model's performance with different adversarial budgets against its clean performance. During the model training phrase, the node feature will be used if it is available on that dataset. Otherwise, the one-hot node id feature will be used as node features for VGAE. The complete performance of all models against RF and EA is shown in Table~\ref{tab:robust} and the robustness towards DICE is presented as Figure~\ref{fig:robust}.

From Table~\ref{tab:robust} and Figure~\ref{fig:robust}, we have the following observations. (1) Generative models, e.g., VGAE and \method, are more robust compared to most of the discriminative based models. This observation is consistent with the robustness conclusion in prior researches on generative and discriminative methods~\cite{ng2001discriminative}. (2) \method achieves dominating leading positions in the relative performance degradation percentage, while keeping leading in the absolute metric values on most of datasets. This phenomenon demonstrates the robustness of \method. (3) Although \method does not shown its steady leading position on some datasets like citeseer, but its performance degradation is relatively much smaller than the other baselines. And we have reason to believe that \method will be more robust when face stronger perturbations. 

%% file: conclusion.tex
% \section{Conclusion and Discussion}
% In this paper, we make the first attempt to adopt the diffusion model to the link prediction problem. With incorporating SGNNs and bayesian decomposition over the generative process of adjacency matrix and node features, we solve the challenges face by prior generative models. With extensive experiments over the model's generalization, robustness and cross-data transferring capability, we successfully demonstrate the advantages of applying generative models toward graph learning tasks. Additionally, through the findings on the exchangeable structure components over different dataset, we show the potential of our proposed framework to be an unified pre-training framework for link prediction in the future. 

% In this paper, we aim to adopt the diffusion model to the link prediction problem. With incorporating SGNNs and Bayesian decomposition over the generative process of graph structure and node features, we solve poor transferring capability and bad memory print on large graphs challenges facing by existing generative models. With extensive experiments over the model's generalization, robustness and cross-data transferring capability, we successfully demonstrate the advantages of applying generative models toward graph learning tasks. Additionally, through the findings on the exchangeable structure components over datasets, we show the potential of our proposed framework to be an unified pre-training framework for link prediction in the future.

In this paper, we aim to adopt the diffusion model to the link prediction problem. With extensive experiments over the model's generalization, robustness and cross-data transfer capability, we successfully demonstrate the advantages of applying generative models toward graph learning tasks. Additionally, through the findings on the exchangeable structure components over datasets, we show the potential of our proposed framework to be an unified pre-training framework for link prediction in the future.

% With incorporating SGNNs and Bayesian decomposition over the generative process of graph structure and node features, we solve poor transferring capability and bad memory print on large graphs challenges facing by existing generative models.

%% file: appendix.tex
\newpage
\section{Appendix}

\subsection{Algorithm Pseudo Code} \label{app:pseudo_code}

The entire process of \method{} is shown in Algorithm~\ref{alg:one}. We estimate the likelihood scores for the structure and features simultaneously from lines 2 to 9 and 10 to 16, respectively. The two components are then fused together on line 17. Lastly, the sample's final connection probability is estimated on line 18.

\RestyleAlgo{ruled}

%% This is needed if you want to add comments in
%% your algorithm with \Comment
\SetKwComment{Comment}{/* }{ */}

\begin{algorithm}[!btph]
\caption{Sub-graph Based Diffusion Model (\method)}\label{alg:one}
\KwIn{Sub-graph $G=({\bf A}, {\bf X})$, connection condition inputs $y_c\in\{0,1\}$, structure diffusion model $\phi_{\theta}$, feature diffusion model $\epsilon_{\theta}$, fusion parameter set $\{\eta_1,\eta_2,\delta\}$, number of steps of structure diffusion model $N_\phi$, number of steps of feature diffusion model $N_\epsilon$.}
Initialize $\mathrm{StructureScore}[y_c]=\mathrm{list()}$ and $\mathrm{FeatureScore}[y_c]=\mathrm{list()}$ for each $y_c$\;
\For{step $t\leftarrow N_\phi$ \KwTo $1$}{
    prepare $\mathbf{X}^{\prime(0)}$ with labeling tricks on $\mathbf{A}^{(0)}$\;
    sample $G^{(t)}$ with $q(G^{(t)}|G^{(0)}) = (\mathbf{A}^{(0)}\Bar{\mathbf{Q}}_A^t,\mathbf{X}^{{\prime}(0)}\Bar{\mathbf{Q}}_X^t)$\;
    \For{conditioning $y_c\leftarrow0$ \KwTo $1$}{
        $\mathrm{StructureScore}[y_c]\mathrm{.append}($\\
        $D_{KL}[q(G^{t-1}|G^t,G)||p_{\theta}(G^{t-1}|G^t,y_c))$
    }
}
\For{step $t\leftarrow N_\epsilon$ \KwTo $1$}{
    sample $\epsilon \sim \mathcal{N}(0, I)$\;
    $\mathbf{X}^{(t)} = \sqrt{\Bar{\alpha}_t}\mathbf{X}^{(0)} + \sqrt{1-\Bar{\alpha}_t}\epsilon$\; 
    \For{conditioning $y_c\leftarrow0$ \KwTo $1$}{
        $\mathrm{FeatureScore}[y_c]\mathrm{.append}(||\epsilon - \epsilon_{\theta}(\mathbf{X}^{(t)},\mathbf{A}^{(0)},y_c)||^2)$\\
    }
}
\KwCal{$\log P(\mathbf{A}|y_c) = \mathrm{mean}(\mathrm{StructureScore}[y_c])$; $\log P(\mathbf{X}|\mathbf{A}, y_c) = \mathrm{mean}(\mathrm{FeatureScore}[y_c])$; $\log P(\mathbf{A},\mathbf{X}|y_c) = \eta_1 \cdot \log P(\mathbf{X}|\mathbf{A},y) + \eta_2 \cdot \log P(\mathbf{A}|y) + \delta$\;
\KwRet $\underset{y_c\in\{0,1\}}{\argmin}(\mathrm{softmax}(\log P({\bf A}, {\bf X}|y_c))$
}

\end{algorithm}

\subsection{Dataset Details} \label{app:dataset_deatil}

The details about each dataset are shown in Tabel~\ref{tab:data_stats}. Following prior works~\cite{kipf2016variational, zhang2018link}, we split the existing links in each graph into train/valid/test with the percentages 80\%/5\%/15\%. For evaluation, we randomly sample the same amount of unconnected node pairs as the negative samples. The evaluation metrics used in our experiment are AUC, Average Precision(AP) and Hit@100. 

\begin{table}[!btph]
\centering
\caption{Detailed statistical information about each dataset.}
\label{tab:data_stats}
\renewcommand{\arraystretch}{1.2}
\resizebox{0.75\textwidth}{!}{
\begin{tabular}{@{}cccccc@{}}
\toprule
Data & Domain & \begin{tabular}[c]{@{}c@{}}Node\\ Number\end{tabular} & \begin{tabular}[c]{@{}c@{}}Edge\\ Number\end{tabular} & \begin{tabular}[c]{@{}c@{}}Average\\ Node Degree\end{tabular} & \begin{tabular}[c]{@{}c@{}}Node\\ Feature / Label\end{tabular} \\ \midrule
Cora & Citation & 2,708 & 10,556 & 3.89 & \ding{52} \\
Citeseer & Citation & 3,327 & 9,228 & 2.77 & \ding{52} \\
Pubmed & Citation & 19,717 & 88,651 & 4.49 & \ding{52} \\
Router & Transporation & 5,022 & 12,516 & 2.49 & \ding{56} \\
USAir & Transporation & 332 & 4,252 & 12.81 & \ding{56} \\
NS & Collaboration & 1,589 & 5,484 & 3.45 & \ding{56} \\ \bottomrule
\end{tabular}}
% \vspace{-1.5cm}
\end{table}

\subsection{Implement Details} 
\label{app:implement_detail}
The implementation and hyper-parameter settings of the two baseline models follow prior works~\cite{kipf2016variational,zhang2018link}. The implementation of our structure diffusion model follows the prior work~\cite{vignac2022digress} and the feature diffusion model is implemented with a multi-layer GCNs. During the enclosing graph generation process, we incorporate the neighbor sampling trick~\cite{hamilton2017inductive} to avoid the graph size becoming extremely large when it encounters some hub nodes. To add DRNL into the structure diffusion process, we treat extracted structure labels as categorical variables and use the sum of node and feature cross-entropy loss to train the structure denoising model. We perform grid search over the hyper parameters of our score and feature diffusion models. The best parameter of each components for each dataset is shown in Table.~\ref{tab:hyper}.

\begin{table}[]
\centering
\caption{Hyper-parameter setting of structure and feature diffusion models.}
\label{tab:hyper}
\renewcommand{\arraystretch}{1.2}
\resizebox{0.8\textwidth}{!}{
\begin{tabular}{@{}|cccccccc|@{}}
\toprule
\multicolumn{1}{|c|}{Name} & \multicolumn{1}{c|}{Symbol} & \multicolumn{1}{c|}{Cora} & \multicolumn{1}{c|}{Citeseer} & \multicolumn{1}{c|}{Pubmed} & \multicolumn{1}{c|}{Router} & \multicolumn{1}{c|}{USAir} & NS \\ \midrule
\multicolumn{8}{|c|}{Sturcture Diffusion} \\ \midrule
\multicolumn{1}{|c|}{\begin{tabular}[c]{@{}c@{}}Hop number of subgraph \\ enclosing the link\end{tabular}} & \multicolumn{1}{c|}{k} & \multicolumn{1}{c|}{1} & \multicolumn{1}{c|}{1} & \multicolumn{1}{c|}{1} & \multicolumn{1}{c|}{1} & \multicolumn{1}{c|}{1} & 1 \\ \midrule
\multicolumn{1}{|c|}{\begin{tabular}[c]{@{}c@{}}Maximum node number \\ for each hop's sampling\end{tabular}} & \multicolumn{1}{c|}{ns} & \multicolumn{1}{c|}{-1} & \multicolumn{1}{c|}{20} & \multicolumn{1}{c|}{20} & \multicolumn{1}{c|}{10} & \multicolumn{1}{c|}{40} & 5 \\ \midrule
\multicolumn{1}{|c|}{\begin{tabular}[c]{@{}c@{}}Attention  hidden neuron \\ number of each layer for\\  node representation\end{tabular}} & \multicolumn{1}{c|}{ha\_x} & \multicolumn{6}{c|}{256} \\ \midrule
\multicolumn{1}{|c|}{\begin{tabular}[c]{@{}c@{}}Attention  hidden neuron \\ number of each layer for \\ edge representation\end{tabular}} & \multicolumn{1}{c|}{ha\_e} & \multicolumn{6}{c|}{64} \\ \midrule
\multicolumn{1}{|c|}{\begin{tabular}[c]{@{}c@{}}Attention  hidden neuron \\ number of each layer for \\ global condition representation\end{tabular}} & \multicolumn{1}{c|}{ha\_y} & \multicolumn{6}{c|}{64} \\ \midrule
\multicolumn{1}{|c|}{\begin{tabular}[c]{@{}c@{}}MLP   hidden neuron \\ number of each layer for \\ node representation\end{tabular}} & \multicolumn{1}{c|}{hm\_x} & \multicolumn{6}{c|}{256} \\ \midrule
\multicolumn{1}{|c|}{\begin{tabular}[c]{@{}c@{}}Attention   hidden neuron \\ number of each layer for \\ edge representation\end{tabular}} & \multicolumn{1}{c|}{hm\_e} & \multicolumn{6}{c|}{128} \\ \midrule
\multicolumn{1}{|c|}{\begin{tabular}[c]{@{}c@{}}Attention   hidden neuron \\ number of each layer for \\ global condition representation\end{tabular}} & \multicolumn{1}{c|}{hm\_y} & \multicolumn{6}{c|}{128} \\ \midrule
\multicolumn{1}{|c|}{Head  number of attention} & \multicolumn{1}{c|}{head} & \multicolumn{6}{c|}{8} \\ \midrule
\multicolumn{1}{|c|}{Number   of transformer layer} & \multicolumn{1}{c|}{l\_t} & \multicolumn{6}{c|}{2} \\ \midrule
\multicolumn{1}{|c|}{Number   of diffusion steps} & \multicolumn{1}{c|}{ds\_t} & \multicolumn{1}{c|}{20} & \multicolumn{1}{c|}{20} & \multicolumn{1}{c|}{10} & \multicolumn{1}{c|}{5} & \multicolumn{1}{c|}{10} & 20 \\ \midrule
\multicolumn{8}{|c|}{Feature   Diffusion} \\ \midrule
\multicolumn{1}{|c|}{Hidden  neruon of GCNs} & \multicolumn{1}{c|}{h\_g} & \multicolumn{1}{c|}{256} & \multicolumn{1}{c|}{256} & \multicolumn{1}{c|}{64} & \multicolumn{3}{c|}{\multirow{3}{*}{N/A}} \\ \cmidrule(r){1-5}
\multicolumn{1}{|c|}{Number   of GCN layers} & \multicolumn{1}{c|}{l\_g} & \multicolumn{1}{c|}{2} & \multicolumn{1}{c|}{2} & \multicolumn{1}{c|}{2} & \multicolumn{3}{c|}{} \\ \cmidrule(r){1-5}
\multicolumn{1}{|c|}{Number   of diffusion steps} & \multicolumn{1}{c|}{ds\_g} & \multicolumn{1}{c|}{100} & \multicolumn{1}{c|}{100} & \multicolumn{1}{c|}{50} & \multicolumn{3}{c|}{} \\ \bottomrule
\end{tabular}}
\end{table}

\subsection{Cross-dataset Performance Details} 
\label{app:crossperform_detail}

We explore the transferability of models by setting each of the six graph as the training dataset and testing the trained model on the other five and itself under the zero-shot scenario. Specicially, detailed performances of seven baseline models and \method on AUC, AP and Hit@100 are presented in Table.~\ref{tab:detail_auc}, Table.~\ref{tab:detail_ap} and Table.~\ref{tab:detail_hit}. For each metric, we run experiment for 10 times and the mean value and standard deviation are reported in the format of mean ± std\%. 

% Please add the following required packages to your document preamble:
% \usepackage{booktabs}
% \usepackage{multirow}
\begin{table}[]
\centering
\caption{Area under the curve (AUC) of different models in cross-data transferability experiment.}
\label{tab:detail_auc}
\resizebox{0.9\textwidth}{!}{
\begin{tabular}{@{}c|c|ccccccc@{}}
\toprule
Target & Source & GCN & GAT & SAGE & NeoGNN & VGAE & SEAL & SGDiff \\ \midrule
\multirow{6}{*}{Cora} & Cora & 90.49 ± 0.59 & 89.85 ± 0.97 & 90.28 ± 0.84 & 92.01 ± 0.61 & 88.98 ± 1.09 & 91.74 ± 0.91 & 90.21 ± 2.21 \\
 & Citeseer & 71.99 ± 3.08 & 80.35 ± 1.47 & 77.29 ± 1.62 & 84.05 ± 1.72 & 67.16 ± 3.87 & 88.11 ± 1.73 & 90.09 ± 0.81 \\
 & Pubmed & 75.14 ± 4.31 & 77.12 ± 0.75 & 79.08 ± 1.29 & 84.49 ± 2.72 & 67.68 ± 3.80 & 88.36 ± 0.52 & 90.73 ± 2.03 \\
 & Router & 72.70 ± 1.42 & 72.18 ± 1.19 & 69.71 ± 1.48 & 82.44 ± 1.51 & 60.69 ± 3.17 & 84.42 ± 1.65 & 86.17 ± 2.31 \\
 & NS & 50.42 ± 4.60 & 66.20 ± 1.11 & 64.61 ± 1.41 & 75.60 ± 1.45 & 59.07 ± 1.58 & 84.08 ± 2.39 & 84.65 ± 1.91 \\
 & USAir & 62.36 ± 2.93 & 62.85 ± 1.88 & 59.01 ± 1.26 & 77.40 ± 2.77 & 59.87 ± 1.35 & 72.57 ± 3.80 & 79.74 ± 5.07 \\ \midrule
\multirow{6}{*}{Citeseer} & Citeseer & 89.64 ± 1.11 & 88.90 ± 1.62 & 89.35 ± 1.28 & 90.60 ± 1.01 & 88.17 ± 0.80 & 89.37 ± 0.99 & 89.36 ± 2.16 \\
 & Cora & 80.12 ± 1.87 & 79.55 ± 1.45 & 79.43 ± 2.64 & 82.82 ± 1.43 & 67.93 ± 2.58 & 89.14 ± 1.04 & 87.81 ± 2.39 \\
 & Pubmed & 75.94 ± 4.77 & 78.04 ± 1.92 & 79.18 ± 1.92 & 81.11 ± 3.02 & 68.34 ± 2.73 & 80.78 ± 1.31 & 91.64 ± 1.77 \\
 & Router & 62.68 ± 1.59 & 69.02 ± 3.04 & 66.06 ± 1.94 & 75.84 ± 2.27 & 58.79 ± 2.73 & 82.45 ± 2.04 & 87.49 ± 2.96 \\
 & NS & 57.36 ± 3.88 & 66.37 ± 2.27 & 64.79 ± 2.34 & 67.45 ± 2.76 & 57.74 ± 2.20 & 86.53 ± 1.16 & 82.93 ± 2.09 \\
 & USAir & 60.98 ± 2.27 & 63.24 ± 2.74 & 59.24 ± 1.91 & 70.45 ± 3.38 & 56.62 ± 1.16 & 70.58 ± 3.15 & 78.13 ± 5.75 \\ \midrule
\multirow{6}{*}{Pubmed} & Pubmed & 96.01 ± 0.30 & 93.07 ± 0.37 & 96.17 ± 0.20 & 96.50 ± 0.32 & 95.37 ± 0.19 & 97.36 ± 0.18 & 95.97 ± 0.75 \\
 & Cora & 89.54 ± 1.96 & 83.13 ± 0.95 & 86.70 ± 1.26 & 87.24 ± 1.49 & 78.10 ± 2.94 & 87.34 ± 2.19 & 90.74 ± 2.17 \\
 & Citeseer & 78.05 ± 5.57 & 82.29 ± 2.45 & 85.50 ± 1.04 & 83.25 ± 1.38 & 77.78 ± 3.73 & 79.54 ± 3.27 & 94.47 ± 1.48 \\
 & Router & 75.93 ± 0.58 & 68.19 ± 1.21 & 78.76 ± 0.54 & 83.61 ± 6.86 & 57.84 ± 4.50 & 87.06 ± 2.36 & 85.59 ± 3.48 \\
 & NS & 33.15 ± 4.98 & 58.91 ± 0.79 & 66.91 ± 0.93 & 67.22 ± 0.58 & 52.27 ± 0.65 & 78.06 ± 7.26 & 88.76 ± 3.73 \\
 & USAir & 57.39 ± 3.76 & 58.34 ± 1.37 & 63.21 ± 0.88 & 77.04 ± 2.09 & 53.71 ± 0.92 & 50.77 ± 7.42 & 89.13 ± 2.39 \\ \midrule
\multirow{6}{*}{Router} & Router & 84.05 ± 1.03 & 64.33 ± 2.32 & 75.23 ± 0.97 & 70.09 ± 4.99 & 63.39 ± 2.35 & 95.90 ± 0.27 & 94.76 ± 0.69 \\
 & Cora & 60.09 ± 3.61 & 49.57 ± 1.94 & 65.65 ± 1.52 & 45.48 ± 3.37 & 53.21 ± 1.61 & 77.99 ± 3.03 & 88.80 ± 2.24 \\
 & Citeseer & 38.67 ± 4.50 & 45.66 ± 1.63 & 64.93 ± 1.12 & 37.46 ± 2.21 & 50.32 ± 0.96 & 66.34 ± 6.74 & 91.90 ± 1.68 \\
 & Pubmed & 70.05 ± 1.16 & 57.13 ± 1.04 & 69.89 ± 1.08 & 69.96 ± 1.24 & 48.15 ± 1.40 & 84.44 ± 1.04 & 94.20 ± 0.91 \\
 & NS & 22.42 ± 3.13 & 42.00 ± 1.62 & 58.89 ± 0.95 & 35.12 ± 1.91 & 52.34 ± 2.18 & 81.99 ± 4.12 & 82.95 ± 8.46 \\
 & USAir & 59.23 ± 4.96 & 43.91 ± 2.18 & 58.11 ± 2.08 & 62.11 ± 5.49 & 56.81 ± 2.66 & 64.03 ± 7.81 & 79.09 ± 5.71 \\ \midrule
\multirow{6}{*}{NS} & NS & 89.79 ± 1.98 & 90.97 ± 1.45 & 91.75 ± 1.09 & 88.87 ± 1.47 & 93.32 ± 0.90 & 98.28 ± 0.35 & 97.47 ± 0.57 \\
 & Cora & 86.88 ± 1.33 & 90.26 ± 1.18 & 87.28 ± 1.65 & 90.75 ± 1.62 & 77.60 ± 1.60 & 97.59 ± 0.42 & 92.16 ± 4.30 \\
 & Citeseer & 87.58 ± 1.16 & 88.28 ± 0.84 & 86.38 ± 1.76 & 89.51 ± 1.64 & 78.74 ± 2.77 & 97.07 ± 0.57 & 95.46 ± 2.66 \\
 & Pubmed & 90.52 ± 1.24 & 90.99 ± 1.58 & 90.16 ± 1.37 & 91.20 ± 1.48 & 84.41 ± 1.62 & 92.62 ± 1.15 & 95.02 ± 1.16 \\
 & Router & 76.78 ± 3.02 & 90.18 ± 1.42 & 84.92 ± 2.44 & 92.44 ± 1.11 & 76.58 ± 5.54 & 89.01 ± 2.84 & 89.15 ± 2.57 \\
 & USAir & 78.54 ± 1.42 & 85.21 ± 1.61 & 74.37 ± 1.68 & 88.10 ± 2.67 & 73.66 ± 2.07 & 94.59 ± 1.77 & 80.43 ± 5.64 \\ \midrule
\multirow{6}{*}{USAir} & USAir & 93.75 ± 1.64 & 95.34 ± 1.10 & 94.91 ± 1.07 & 94.15 ± 1.50 & 90.84 ± 1.21 & 97.62 ± 0.55 & 96.25 ± 1.58 \\
 & Cora & 86.38 ± 2.19 & 83.48 ± 2.43 & 85.11 ± 2.37 & 83.25 ± 13.72 & 65.53 ± 5.33 & 90.76 ± 2.43 & 65.90 ± 24.32 \\
 & Citeseer & 86.34 ± 2.46 & 81.83 ± 2.95 & 84.21 ± 2.40 & 62.04 ± 13.81 & 79.33 ± 4.57 & 86.86 ± 4.01 & 81.65 ± 11.75 \\
 & Pubmed & 92.90 ± 1.39 & 87.43 ± 2.88 & 90.53 ± 1.49 & 92.68 ± 1.78 & 88.81 ± 1.67 & 89.49 ± 1.18 & 84.84 ± 12.23 \\
 & Router & 77.06 ± 3.07 & 86.80 ± 1.30 & 83.11 ± 1.64 & 82.13 ± 17.75 & 70.04 ± 12.63 & 89.04 ± 2.05 & 84.79 ± 3.49 \\
 & NS & 51.92 ± 9.13 & 81.07 ± 1.77 & 76.82 ± 3.07 & 48.71 ± 6.89 & 59.31 ± 2.54 & 90.34 ± 1.96 & 91.11 ± 2.95 \\ \bottomrule
\end{tabular}}
\end{table}

% Please add the following required packages to your document preamble:
% \usepackage{booktabs}
% \usepackage{multirow}
\begin{table}[]
\centering
\caption{Average precision (AP) of different models in cross-data transferability experiment.}
\label{tab:detail_ap}
\resizebox{0.9\textwidth}{!}{
\begin{tabular}{@{}c|c|ccccccc@{}}
\toprule
Target & Source & GCN & GAT & SAGE & NeoGNN & VGAE & SEAL & SGDiff \\ \midrule
\multirow{6}{*}{Cora} & Cora & 92.16 ± 0.40 & 91.14 ± 0.69 & 91.43 ± 0.87 & 93.39 ± 0.50 & 90.81 ± 0.83 & 92.85 ± 0.64 & 90.00 ± 2.41 \\
 & Citeseer & 76.79 ± 2.49 & 82.48 ± 1.22 & 78.37 ± 1.70 & 88.17 ± 1.05 & 67.47 ± 3.82 & 90.21 ± 1.40 & 91.08 ± 1.11 \\
 & Pubmed & 77.79 ± 3.86 & 80.26 ± 0.99 & 81.15 ± 1.20 & 88.79 ± 1.56 & 67.43 ± 4.72 & 90.86 ± 0.47 & 91.92 ± 1.86 \\
 & Router & 74.05 ± 1.45 & 77.65 ± 0.87 & 70.89 ± 1.27 & 86.74 ± 1.75 & 61.39 ± 4.15 & 86.90 ± 1.30 & 85.85 ± 3.17 \\
 & NS & 58.42 ± 4.08 & 71.68 ± 1.19 & 65.44 ± 0.92 & 82.44 ± 1.63 & 58.96 ± 1.71 & 86.00 ± 1.47 & 86.15 ± 1.85 \\
 & USAir & 62.80 ± 2.40 & 66.42 ± 2.42 & 57.48 ± 1.38 & 78.76 ± 3.95 & 59.32 ± 1.27 & 78.18 ± 3.58 & 82.55 ± 3.98 \\ \midrule
\multirow{6}{*}{Citeseer} & Citeseer & 91.54 ± 0.94 & 90.93 ± 1.36 & 91.23 ± 1.12 & 92.41 ± 0.99 & 90.19 ± 0.85 & 91.62 ± 0.91 & 90.87 ± 1.93 \\
 & Cora & 80.59 ± 2.04 & 81.67 ± 1.96 & 81.80 ± 2.97 & 87.44 ± 0.99 & 67.36 ± 2.22 & 91.28 ± 0.70 & 88.95 ± 2.02 \\
 & Pubmed & 77.94 ± 4.75 & 83.54 ± 1.72 & 82.63 ± 1.52 & 86.62 ± 2.05 & 67.55 ± 3.15 & 86.52 ± 1.24 & 92.46 ± 1.48 \\
 & Router & 64.64 ± 1.14 & 76.56 ± 2.10 & 68.65 ± 1.76 & 81.66 ± 3.05 & 58.43 ± 2.36 & 85.03 ± 1.79 & 86.66 ± 4.02 \\
 & NS & 63.72 ± 3.10 & 74.11 ± 1.92 & 66.03 ± 2.82 & 78.70 ± 1.89 & 57.15 ± 2.19 & 89.43 ± 0.97 & 85.18 ± 2.17 \\
 & USAir & 59.55 ± 1.94 & 69.57 ± 2.31 & 57.79 ± 1.56 & 72.32 ± 3.94 & 55.39 ± 1.27 & 77.47 ± 3.06 & 77.77 ± 6.36 \\ \midrule
\multirow{6}{*}{Pubmed} & Pubmed & 96.28 ± 0.30 & 93.73 ± 0.29 & 96.29 ± 0.24 & 97.04 ± 0.25 & 96.08 ± 0.20 & 97.38 ± 0.17 & 95.91 ± 1.14 \\
 & Cora & 90.84 ± 1.38 & 83.57 ± 1.26 & 86.82 ± 1.66 & 90.96 ± 0.95 & 78.87 ± 3.33 & 88.59 ± 1.64 & 90.77 ± 2.18 \\
 & Citeseer & 83.65 ± 4.32 & 82.88 ± 3.50 & 85.39 ± 1.40 & 88.27 ± 0.91 & 78.57 ± 4.34 & 83.58 ± 2.56 & 94.71 ± 1.48 \\
 & Router & 75.35 ± 0.67 & 74.48 ± 1.34 & 76.68 ± 0.77 & 87.68 ± 5.16 & 58.19 ± 6.46 & 88.05 ± 2.17 & 83.14 ± 4.23 \\
 & NS & 41.07 ± 3.42 & 63.73 ± 1.33 & 63.31 ± 1.19 & 76.37 ± 2.95 & 50.45 ± 0.56 & 79.39 ± 4.81 & 89.59 ± 2.35 \\
 & USAir & 59.05 ± 2.32 & 61.54 ± 2.04 & 58.71 ± 0.90 & 79.47 ± 1.63 & 54.09 ± 0.98 & 56.52 ± 5.29 & 91.25 ± 1.56 \\ \midrule
\multirow{6}{*}{Router} & Router & 85.60 ± 1.01 & 72.16 ± 2.20 & 78.39 ± 1.60 & 77.53 ± 3.67 & 67.73 ± 1.85 & 95.78 ± 0.29 & 94.60 ± 1.00 \\
 & Cora & 65.25 ± 3.35 & 56.96 ± 2.33 & 65.97 ± 1.52 & 61.65 ± 2.47 & 55.76 ± 1.28 & 81.42 ± 2.26 & 89.70 ± 1.49 \\
 & Citeseer & 46.31 ± 3.28 & 55.85 ± 1.55 & 64.93 ± 1.57 & 53.99 ± 3.46 & 53.11 ± 0.89 & 72.38 ± 5.33 & 92.01 ± 1.86 \\
 & Pubmed & 75.82 ± 1.15 & 64.60 ± 1.26 & 71.79 ± 1.18 & 77.18 ± 1.11 & 52.72 ± 1.78 & 88.63 ± 0.82 & 93.95 ± 0.90 \\
 & NS & 40.44 ± 2.00 & 50.49 ± 2.10 & 57.38 ± 1.15 & 54.00 ± 4.46 & 54.25 ± 1.82 & 82.96 ± 3.28 & 83.05 ± 6.31 \\
 & USAir & 63.68 ± 3.67 & 50.28 ± 1.61 & 56.08 ± 1.77 & 68.92 ± 5.06 & 59.78 ± 2.23 & 66.72 ± 6.39 & 84.40 ± 4.73 \\ \midrule
\multirow{6}{*}{NS} & NS & 93.27 ± 1.72 & 93.85 ± 0.74 & 94.00 ± 0.88 & 93.77 ± 0.83 & 94.74 ± 0.74 & 98.53 ± 0.29 & 97.76 ± 0.60 \\
 & Cora & 87.92 ± 1.93 & 93.85 ± 0.81 & 89.58 ± 1.35 & 94.59 ± 0.94 & 74.49 ± 1.68 & 98.00 ± 0.39 & 94.55 ± 2.70 \\
 & Citeseer & 91.02 ± 0.93 & 92.32 ± 0.59 & 88.77 ± 1.62 & 94.07 ± 0.92 & 75.98 ± 3.08 & 97.65 ± 0.37 & 95.11 ± 3.32 \\
 & Pubmed & 93.61 ± 0.81 & 94.08 ± 0.89 & 92.32 ± 0.96 & 94.57 ± 0.96 & 82.18 ± 2.09 & 94.85 ± 1.09 & 95.79 ± 1.27 \\
 & Router & 74.83 ± 3.48 & 93.66 ± 0.89 & 87.33 ± 1.95 & 94.99 ± 0.66 & 73.91 ± 5.63 & 90.36 ± 2.40 & 87.93 ± 3.75 \\
 & USAir & 74.33 ± 1.56 & 89.54 ± 1.25 & 72.80 ± 1.82 & 89.42 ± 3.48 & 68.82 ± 2.75 & 96.15 ± 1.18 & 78.21 ± 7.62 \\ \midrule
\multirow{6}{*}{USAir} & USAir & 95.24 ± 1.12 & 95.89 ± 0.82 & 94.55 ± 1.04 & 95.84 ± 1.09 & 93.00 ± 0.98 & 97.51 ± 0.86 & 96.84 ± 1.39 \\
 & Cora & 85.73 ± 2.55 & 81.46 ± 4.03 & 83.24 ± 3.65 & 84.95 ± 12.86 & 59.94 ± 6.20 & 91.07 ± 3.55 & 72.79 ± 17.80 \\
 & Citeseer & 86.89 ± 2.64 & 79.64 ± 3.26 & 80.82 ± 4.13 & 65.22 ± 13.01 & 79.95 ± 5.95 & 86.93 ± 5.12 & 84.36 ± 8.04 \\
 & Pubmed & 94.22 ± 1.35 & 85.22 ± 4.18 & 90.03 ± 1.57 & 94.08 ± 1.81 & 91.27 ± 1.40 & 90.83 ± 1.37 & 85.84 ± 8.85 \\
 & Router & 74.80 ± 3.13 & 86.09 ± 2.05 & 80.22 ± 2.01 & 83.47 ± 15.26 & 69.21 ± 15.81 & 87.97 ± 2.52 & 81.08 ± 4.30 \\
 & NS & 55.78 ± 6.96 & 80.01 ± 1.71 & 72.94 ± 4.36 & 50.48 ± 4.39 & 53.57 ± 2.24 & 92.67 ± 2.46 & 90.24 ± 4.21 \\ \bottomrule
\end{tabular}}
\end{table}

% Please add the following required packages to your document preamble:
% \usepackage{booktabs}
% \usepackage{multirow}
\begin{table}[]
\centering
\caption{Hit@100 of different models in cross-data transferability experiment.}
\label{tab:detail_hit}
\resizebox{0.9\textwidth}{!}{
\begin{tabular}{@{}c|c|ccccccc@{}}
\toprule
Target & Source & GCN & GAT & SAGE & NeoGNN & VGAE & SEAL & SGDiff \\ \midrule
\multirow{6}{*}{Cora} & Cora & 85.97 ± 1.06 & 84.94 ± 1.78 & 84.43 ± 1.56 & 87.36 ± 2.02 & 81.87 ± 1.80 & 87.56 ± 2.24 & 83.68 ± 4.48 \\
 & Citeseer & 56.57 ± 3.22 & 66.23 ± 2.66 & 58.34 ± 2.97 & 76.13 ± 2.60 & 41.64 ± 6.27 & 81.86 ± 2.75 & 84.49 ± 1.64 \\
 & Pubmed & 59.31 ± 6.58 & 61.78 ± 1.86 & 64.82 ± 2.69 & 76.59 ± 3.75 & 41.70 ± 6.66 & 82.26 ± 2.34 & 85.76 ± 3.76 \\
 & Router & 50.81 ± 3.66 & 56.53 ± 1.67 & 48.61 ± 3.53 & 72.88 ± 2.98 & 33.03 ± 6.65 & 74.72 ± 3.72 & 76.48 ± 5.75 \\
 & NS & 28.19 ± 5.61 & 44.88 ± 3.04 & 39.54 ± 3.08 & 69.92 ± 2.06 & 28.56 ± 2.64 & 76.31 ± 5.02 & 72.69 ± 3.30 \\
 & USAir & 35.85 ± 3.76 & 39.08 ± 3.82 & 29.42 ± 2.50 & 66.90 ± 6.91 & 31.18 ± 3.48 & 60.26 ± 5.85 & 70.58 ± 8.60 \\ \midrule
\multirow{6}{*}{Citeseer} & Citeseer & 85.39 ± 1.93 & 83.20 ± 2.47 & 84.44 ± 1.54 & 85.30 ± 1.10 & 81.87 ± 1.23 & 85.99 ± 2.44 & 84.34 ± 4.09 \\
 & Cora & 68.89 ± 3.02 & 69.90 ± 2.67 & 67.68 ± 5.43 & 74.10 ± 2.25 & 44.70 ± 4.69 & 86.88 ± 2.01 & 82.55 ± 5.19 \\
 & Pubmed & 63.97 ± 7.51 & 68.11 ± 3.33 & 69.66 ± 3.13 & 72.32 ± 3.89 & 46.42 ± 6.43 & 72.99 ± 2.30 & 88.44 ± 3.00 \\
 & Router & 40.31 ± 3.01 & 56.82 ± 3.78 & 49.03 ± 3.13 & 65.08 ± 3.69 & 32.06 ± 3.95 & 71.84 ± 5.42 & 83.08 ± 4.04 \\
 & NS & 39.98 ± 5.08 & 53.45 ± 3.61 & 44.70 ± 4.47 & 59.42 ± 2.78 & 28.27 ± 3.24 & 81.10 ± 2.22 & 70.49 ± 5.02 \\
 & USAir & 35.19 ± 3.45 & 47.15 ± 3.52 & 34.15 ± 3.05 & 59.89 ± 7.71 & 26.99 ± 3.24 & 58.01 ± 8.41 & 71.59 ± 8.65 \\ \midrule
\multirow{6}{*}{Pubmed} & Pubmed & 73.75 ± 3.23 & 64.00 ± 2.08 & 73.53 ± 1.59 & 78.90 ± 1.85 & 74.22 ± 1.48 & 75.21 ± 1.43 & 68.77 ± 8.79 \\
 & Cora & 50.94 ± 4.53 & 28.90 ± 3.76 & 35.51 ± 4.32 & 60.68 ± 5.67 & 22.74 ± 5.04 & 44.20 ± 5.27 & 44.38 ± 10.69 \\
 & Citeseer & 44.58 ± 7.02 & 29.59 ± 8.35 & 30.35 ± 4.80 & 52.23 ± 9.10 & 22.81 ± 6.28 & 44.32 ± 5.29 & 65.81 ± 8.92 \\
 & Router & 17.70 ± 1.14 & 25.68 ± 3.13 & 13.59 ± 2.02 & 51.12 ± 11.81 & 5.61 ± 3.48 & 42.64 ± 5.97 & 24.02 ± 10.25 \\
 & NS & 1.36 ± 0.71 & 11.25 ± 1.90 & 3.91 ± 0.83 & 28.64 ± 20.89 & 2.50 ± 0.37 & 22.92 ± 10.61 & 45.86 ± 6.43 \\
 & USAir & 2.95 ± 3.62 & 8.07 ± 2.40 & 2.10 ± 0.30 & 18.53 ± 9.65 & 4.86 ± 0.53 & 5.26 ± 2.38 & 53.75 ± 8.30 \\ \midrule
\multirow{6}{*}{Router} & Router & 68.28 ± 2.66 & 44.53 ± 3.30 & 57.77 ± 1.76 & 55.55 ± 5.69 & 36.55 ± 3.89 & 93.11 ± 1.78 & 89.47 ± 3.41 \\
 & Cora & 38.95 ± 4.66 & 25.20 ± 2.73 & 40.67 ± 2.80 & 34.71 ± 3.29 & 23.93 ± 2.36 & 58.70 ± 7.77 & 78.41 ± 4.29 \\
 & Citeseer & 14.82 ± 3.68 & 23.09 ± 1.84 & 39.26 ± 2.31 & 29.73 ± 2.30 & 20.71 ± 1.68 & 44.98 ± 10.47 & 82.71 ± 3.27 \\
 & Pubmed & 54.16 ± 2.18 & 32.35 ± 1.77 & 49.49 ± 2.29 & 54.23 ± 2.04 & 23.11 ± 2.84 & 77.06 ± 2.17 & 88.19 ± 5.71 \\
 & NS & 4.76 ± 2.30 & 19.70 ± 1.90 & 26.84 ± 2.49 & 28.02 ± 1.83 & 21.49 ± 2.62 & 65.66 ± 9.07 & 65.79 ± 14.49 \\
 & USAir & 37.04 ± 6.24 & 19.30 ± 1.76 & 24.70 ± 2.98 & 48.32 ± 5.97 & 29.72 ± 3.13 & 37.99 ± 10.19 & 67.46 ± 11.50 \\ \midrule
\multirow{6}{*}{NS} & NS & 88.31 ± 2.16 & 89.02 ± 2.03 & 89.95 ± 1.63 & 87.57 ± 1.83 & 92.77 ± 1.21 & 98.92 ± 0.69 & 98.54 ± 1.28 \\
 & Cora & 88.62 ± 1.83 & 88.79 ± 1.58 & 87.42 ± 2.12 & 89.53 ± 2.35 & 79.89 ± 3.75 & 98.39 ± 0.80 & 91.66 ± 5.77 \\
 & Citeseer & 87.37 ± 0.91 & 86.32 ± 0.84 & 87.14 ± 2.87 & 88.56 ± 1.96 & 81.06 ± 5.13 & 97.30 ± 1.17 & 96.77 ± 2.46 \\
 & Pubmed & 89.06 ± 1.54 & 88.90 ± 1.74 & 87.80 ± 2.05 & 89.27 ± 1.88 & 87.99 ± 1.65 & 93.23 ± 1.68 & 96.26 ± 1.39 \\
 & Router & 75.13 ± 4.47 & 88.57 ± 1.82 & 85.18 ± 2.65 & 92.07 ± 2.40 & 77.03 ± 9.37 & 87.68 ± 6.58 & 94.52 ± 3.24 \\
 & USAir & 83.05 ± 3.09 & 84.78 ± 1.98 & 73.44 ± 3.57 & 91.69 ± 1.95 & 74.14 ± 4.42 & 94.24 ± 2.74 & 85.28 ± 6.62 \\ \midrule
\multirow{6}{*}{USAir} & USAir & 95.77 ± 1.63 & 96.95 ± 1.20 & 97.92 ± 1.16 & 96.05 ± 1.44 & 92.90 ± 1.54 & 99.78 ± 0.39 & 98.38 ± 0.62 \\
 & Cora & 94.21 ± 1.77 & 91.36 ± 1.93 & 92.62 ± 2.75 & 86.18 ± 17.49 & 71.32 ± 7.43 & 94.77 ± 1.45 & 69.26 ± 26.48 \\
 & Citeseer & 91.48 ± 2.84 & 90.62 ± 2.56 & 92.57 ± 2.93 & 58.84 ± 17.22 & 84.17 ± 4.74 & 92.30 ± 3.24 & 86.12 ± 11.80 \\
 & Pubmed & 95.00 ± 1.66 & 93.97 ± 1.21 & 94.98 ± 1.93 & 95.32 ± 1.62 & 91.91 ± 1.69 & 93.13 ± 2.32 & 88.68 ± 15.24 \\
 & Router & 84.39 ± 4.53 & 92.91 ± 1.74 & 90.54 ± 2.14 & 84.01 ± 24.45 & 72.60 ± 15.83 & 96.26 ± 1.49 & 93.37 ± 3.52 \\
 & NS & 49.36 ± 12.58 & 89.01 ± 2.74 & 86.01 ± 3.94 & 39.81 ± 7.77 & 63.12 ± 2.84 & 95.86 ± 1.98 & 96.70 ± 1.74 \\ \bottomrule
\end{tabular}}
\end{table}